\pdfoutput=1

\documentclass[11pt]{article}

\usepackage[]{acl}
\usepackage[para]{footmisc}

\usepackage{times}
\usepackage{latexsym}

\usepackage[T1]{fontenc}

\usepackage[utf8]{inputenc}

\usepackage{microtype}

\usepackage{inconsolata}

\usepackage{graphicx}

\usepackage{amsmath}
\usepackage{amssymb}
\usepackage{booktabs}
\usepackage{authblk} 

\usepackage{color, booktabs,subcaption,amsfonts,dcolumn}
\usepackage{xcolor}
\usepackage{mdframed}
\usepackage{subcaption}
\newmdenv[leftmargin=5pt,rightmargin=5pt,backgroundcolor=gray!20,innertopmargin=5pt,innerbottommargin=5pt]{coloredquote}

\newenvironment{smallermdframed}{
  \begin{coloredquote} \scriptsize 
}{\end{coloredquote}}

\usepackage{amsfonts}
\usepackage{graphicx}

\usepackage{multirow,diagbox,makecell}
\usepackage{tikz}

\usepackage{pifont}
\usepackage{booktabs}

\usepackage{capt-of}

\title{Read Anywhere Pointed:\\ Layout-aware GUI Screen Reading with Tree-of-Lens Grounding
}

\author[1]{Yue Fan}
\author[1]{Lei Ding}
\author[2]{Ching-Chen Kuo}
\author[2]{Shan Jiang}
\author[2]{Yang Zhao}
\author[2]{Xinze Guan}
\author[3]{Jie Yang}
\author[1]{\\Yi Zhang}
\author[1]{Xin Eric Wang}
\affil[1]{University of California, Santa Cruz}
\affil[2]{eBay Inc.}
\affil[3]{Cybever}

\begin{document}
\maketitle
\begin{abstract}

\end{abstract}
Graphical User Interfaces (GUIs) are central to our interaction with digital devices and growing efforts have been made to build models for various GUI understanding tasks. However, these efforts largely overlook an important GUI-referring task: screen reading based on user-indicated points, which we name the Screen Point-and-Read (ScreenPR) task. Currently, this task is predominantly handled by rigid accessible screen reading tools, in great need of new models driven by advancements in Multimodal Large Language Models (MLLMs). In this paper, we propose a Tree-of-Lens (ToL) agent, utilizing a novel ToL grounding mechanism, to address the ScreenPR task. Based on the input point coordinate and the corresponding GUI screenshot, our ToL agent constructs a Hierarchical Layout Tree. Based on the tree, our ToL agent not only comprehends the content of the indicated area but also articulates the layout and spatial relationships between elements. Such layout information is crucial for accurately interpreting information on the screen, distinguishing our ToL agent from other screen reading tools. We also thoroughly evaluate the ToL agent against other baselines on a newly proposed ScreenPR benchmark, which includes GUIs from mobile, web, and operating systems. Last but not least, we test the ToL agent on mobile GUI navigation tasks, demonstrating its utility in identifying incorrect actions along the path of agent execution trajectories. Code and data: \url{https://screen-point-and-read.github.io}. 



\begin{figure}[t]
    \centering
    \includegraphics[width=0.95\columnwidth]{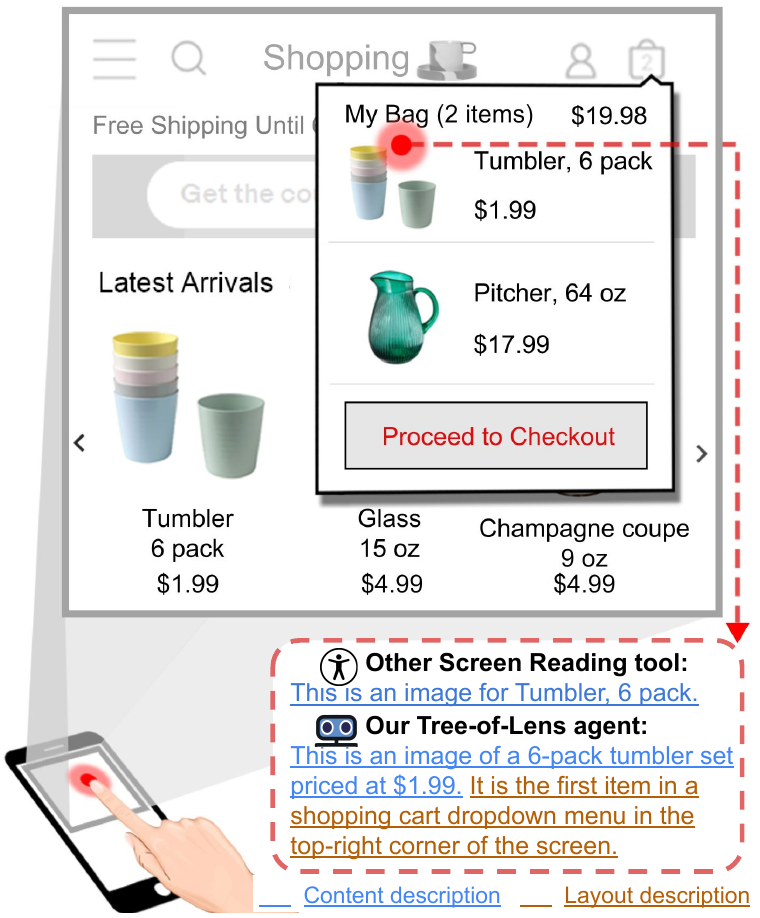}
    \caption{Our ToL agent describes the region on the screenshot indicated by a user's point. Distinguished from other screen reading tools, our ToL agent can output layout-aware descriptions for points anywhere on the screen. }
    \label{fig:1}
\end{figure}

\section{Introduction}
Graphical User Interfaces (GUIs), dominate our digital interactions with visually rich screenshots featuring colors, icons, texts, and spatial layouts. Recognizing that screenshots are more accessible and intuitive for depicting visual cues, there are growing efforts to build AI agents to interpret GUIs visually \cite{shaw2023pixels,deng2023mindweb, cheng2024seeclick, hong2023cogagent, you2024ferret}.
Existing works, however, overlook the point-based screen reading task, where the input is the coordinate of a user's indicated point on the screen with the corresponding screenshot, and the output is a descriptive interpretation of the screen region pointed. This task is critical for accessible technology, providing valuable assistance to users with visual impairments and we refer to such a task as the Screen Point-and-Read (ScreenPR) task. 



Aiming to solve the ScreenPR task, we introduce the Tree-of-Lens (ToL) agent. Taking advantage of the generalizability of the advanced Multimodal Large Language Models (MLLMs), the ToL agent is built for GUI screenshots across domains with a point anywhere on the screen as input. The output from the ToL agent is natural language descriptions for both the content of the indicated region and the related information about the screen layout, as shown in Figure~\ref{fig:1}. Especially, we find that it is critical to articulate the layout between elements in the screenshot for users to gain a comprehensive understanding of the interface and to avoid ambiguity. For example, with only content description, one cannot distinguish the two identical ``Tumbler pack" shown in Figure~\ref{fig:1}. 

Our ToL agent employs a ToL grounding mechanism, constructing a Hierarchical Layout Tree for each input screenshot to depict its underlying structure. The nodes in the tree represent regions of varying scales. We build this tree using an object detection model trained on our newly collected Android Screen Hierarchical Layout (ASHL) dataset, which contains 50k bounding boxes of hierarchical screen regions from Android screenshots. This model enables an automatic extraction of local and global regions, forming the Hierarchical Layout Tree. After constructing the tree, we extract the target path based on the region of interest, producing lenses as visual prompts for the  MLLM, simulating a human-like, gradually refined focus.

To rigorously evaluate our ToL agent against other baselines, we introduce the Screen Point-and-Read (ScreenPR) benchmark with the ScreenPR task that requires the model to generate descriptions based on user-indicated points. This benchmark consists of 650 screenshots from web, mobile, and operating system GUIs, manually annotated with 1,500 target points and regions. With the ScreenPR benchmark, we show that our ToL agent achieves the best performance compared with the baselines of other general and GUI-specialized MLLMs, with over 15\% and 30\% improvements respectively in terms of the accuracy of content and layout descriptions compared with vanilla text-prompting GPT-4o.




In addition, we also test our ToL agent with trajectories from a mobile GUI navigation agent. By applying the ToL agent to describe the point of each action taken in the trajectory, we demonstrate its utility in identifying incorrect actions along the execution path. This capability is pivotal for refining the development of mobile agents, as it provides clear, actionable feedback on navigation decisions. This application not only validates the ToL agent's effectiveness but also opens up possibilities for strengthening the performance of mobile agents with sophisticated, layout-aware GUI understanding capabilities.

In conclusion, the contributions of our paper are as follows:

\begin{itemize}

    \item We develop the Tree-of-Lens (ToL) grounding method and build the ToL agent that generates descriptions that effectively combine both content and layout information of the pointed screen regions. 

    \item We propose the Hierarchical Layout Tree to represent the underlying hierarchical structure of the screenshot. We collect the Android Screen Hierarchical Layout dataset and train a model to build the Hierarchical Layout Tree.

    \item We introduce the Screen Point-and-Read benchmark, which specifically challenges the model to produce accurate descriptions that include both content and layout details.

    \item We rigorously test the ToL agent on the Screen Point-and-Read benchmark and we demonstrate its ability to accurately identify incorrect actions in mobile navigation trajectories.

\end{itemize}

\begin{figure*}
    \centering
    \includegraphics[width=\textwidth]{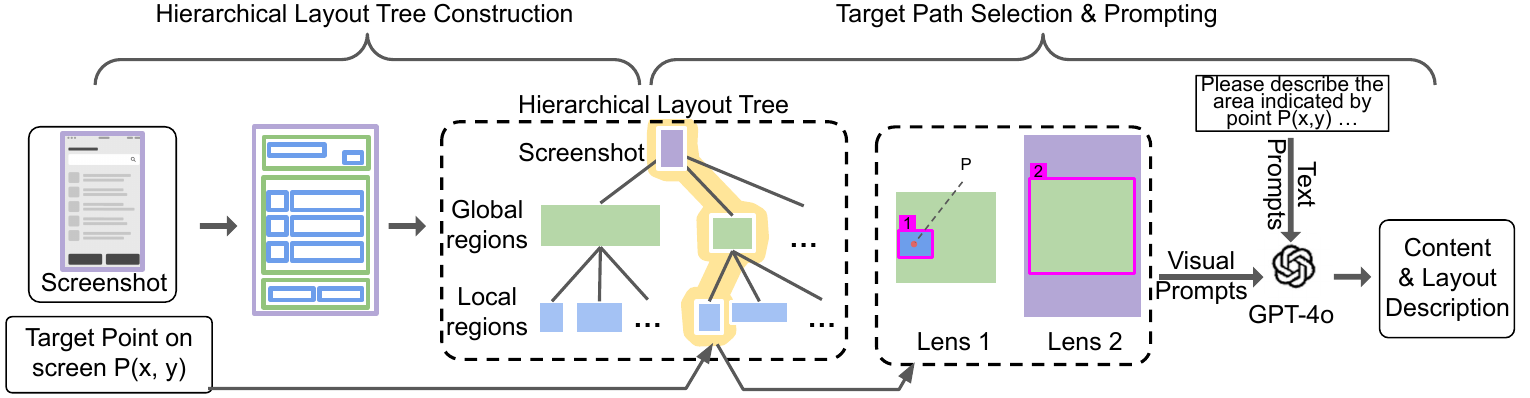}
    \caption{Pipeline of the Tree-of-Lens agent. The Hierarchical Layout Tree is first constructed based on detected global and local regions from the input screenshot. Then, a set of hierarchical lenses with various field widths is generated from the selected target path in the tree and sent as visual prompts to GPT-4o to generate the content and layout descriptions. }
    \label{fig:2}
\end{figure*}

\section{Related Works}



\paragraph{MLLMs for GUI Understanding} 


Recent advancements in Multimodal Large Language Models (MLLMs) have led to a rising series of highly capable generalist models \cite{liu2024visual, chen2023minigptv2, ye2023mplugowl, dai2023instructblip, bai2023qwen, liu2024llavanext}.
Spurred by the strong general capability of MLLMs, recent works have been focused on utilizing and improving MLLMs for GUI tasks. Some focus mainly on a specific GUI task such as GUI navigation \cite{wang2024mobile, yang2023appagent, zheng2023seeact}, GUI referring expression comprehension \cite{cheng2024seeclick}, while some aim at a set of GUI grounding and referring tasks simultaneously, such as CogAgent \cite{hong2023cogagent}, which is Multi-task Fine-tuned on CogVLM \cite{wang2023cogvlm}, and Ferret-UI \cite{you2024ferret}, ensuing Ferret \cite{you2023ferret, zhang2024ferret}, tailored for GUI domain. In our work, we focus on the Screen Point-and-Read (ScreenPR) task, a referring task in which inputs are point locations on the screen with the corresponding screenshots, and develop a specific benchmark to evaluate the model's performance in such tasks. Also, inspired by big challenges that MLLMs are facing in understanding images with complex layout information \cite{fan2024muffin}, we propose the Tree-of-Lens (ToL) agent that generates layout-aware description, which distinguishes from all other GUI agents with GUI referring ability.


\paragraph{Chain-of-Thoughts for Vision} 

The Chain of Thoughts (CoT) \cite{wei2022chain} methodology in natural language processing (NLP) has spurred various inspiring ideas in this direction. \cite{creswell2022selection} applied it to interpretable logical reasoning, \cite{wang2022self} focused on enhancing the "exploring-selecting" mechanism in reasoning path, and \cite{yao2024tree, besta2024graph} brought in more sophisticated patterns for reasoning. To embrace CoT in the vision area, \citet{xi2023chain} uses Chain-of-Look prompting for surgical scene understanding to learn rich semantics from surgical endoscopic videos. In comparison, \citet{chen2024visual} designated an iterative step-by-step reasoning manner for VQA problem, letting MLLMs "See-Think-Confirm" to generate a better rationale for the decided answer. Recently, \citet{shao2024visual} proposed a multi-turn processing pipeline that dynamically focuses on visual inputs and provides interpretable thoughts. Moving along with the direction, we aim to facilitate MLLMs on the GUI referring task by generating a chain of lenses with varying field widths, ranging from fine-grained to coarse views. The chain of lenses demonstrates the hierarchical layout of the GUI, simulating a human-like sequence of attention, improving MLLMs on producing robust and accurate descriptions with the content and layout information regarding the target point input in the accessible screen reading task.


\section{Tree-of-Lens Agent}

\subsection{Task Definition}

We aim to develop an agent to interpret designated regions on GUIs indicated by points, which we refer to as the Screen Point-and-Read (ScreenPR) task. This agent needs to process input consisting of a screenshot \( S_i \) and a point coordinate \( P_i \) on the screen, where the point represents a region of the user's interest. Two target outputs are language descriptions \( \hat{D}^c_i \) for the content of the targeted region and \( \hat{D}^l_i \) for the layout information of the screen related to the targeted region. 

\subsection{Method Overview}

Targeting the ScreenPR task, we propose the Tree-of-Lens (ToL) agent equipped with the ToL grounding mechanism, as shown in Figure \ref{fig:2}. 
Since our agent relies on the pure visual modality of the GUI and leverages the strong visual referring capabilities of MLLMs, such as GPT-4o, it can effectively read anywhere pointed on screens from any domain, including but not limited to web, mobile, and operating systems.

The three distinct design goals distinguish our LoT agent from previous models. First, our LoT agent outputs rich, natural language descriptions rather than fixed template descriptions from accessible screen reading tools such as VoiceOver\footnote{\url{https://support.apple.com/guide/voiceover}}. Second, it can handle any point location on the screen as input by dynamically detecting the pointed region, which might be as small as an icon or a broader area, such as a full-screen window. Last but not least, the descriptions generated by our agent encompass both the content and layout perspectives, detailing the targeted region's function and position within the GUI’s layout. 



\subsection{Hierarchical Layout Tree Construction}
\paragraph{Hierarchical Layout Tree}
The layout of the GUI screenshot usually features an inherent hierarchy decided by GUI source codes. We propose the Hierarchical Layout Tree to represent this hierarchical structure. This tree structure corresponds to each screenshot, with each node representing a square region within the GUI screenshot. Instead of using varied depths of tree structures from the underlying source codes of GUIs, for simplicity, the proposed Hierarchical Layout Tree has a fixed 3-layer depth for every GUI screenshot, with nodes in each layer representing regions of varied hierarchies. The top layer node represents the whole screenshot. The middle layer nodes represent global regions, which encompass larger, comprehensive areas that include multiple related elements, such as panels or groups of controls, forming significant parts of the user interface. The leaf nodes represent local regions, referring to smaller, more specific areas, typically denoting individual interactive elements like buttons, text fields, and icons.

\paragraph{Model and tree construction pipeline}
In order to extract the Hierarchical Layout Tree that depicts the hierarchical layout of screenshots without relying on such structured text modality, we train a GUI region detection model to detect the local and global regions for each GUI screenshot, then construct the Hierarchical Layout Trees accordingly. The GUI region detection model is fine-tuned on the DINO detection model with ResNet-50 backbone \cite{zhang2022dino,MMDetection}. Once the regions are detected, we construct the Hierarchical Layout Tree, where each predicted global region serves as a node connected to one or more leaves representing the predicted local regions. The connection between global and local regions is determined by the highest IoU between the local and global regions, where for a predicted local region $\hat{R_i}$, its parent global region is $\hat{R_j}$, $ j=\textit{ArgMax}_j \textit{IoU}(R_i,R_j)$.

\paragraph{Android Screen Hierarchical Layout dataset} \label{android_screen_hierarchical_layout_dataset}
To train a detection model for the Hierarchical Layout Tree, we build a dataset named the Android Screen Hierarchical Layout (ASHL) dataset, featuring screenshots and bounding boxes labeled as either global or local regions. There are around 52k bounding boxes from 2,180 screenshots, spanning all 18 different Android applications. 

The dataset collection process is fully automated based on the MagicWand simulator \cite{dingenhancing}, where we first sample screenshots along with their corresponding source codes based on pre-recorded actions of humans navigating different Android applications. Then, we extract GUI regions $R_i$ from the source codes with predefined coordinates to become nodes and leaves in a tree structure. The parent-child relationships $\textit{Parent-Child}(R_i, R_j) establish  = 1$ are also inherent directly from the code. Next, we refine the tree by pruning branches corresponding to non-visible regions and merging connected nodes with an Intersection Over Union (IoU) greater than 0.9 to a single node, representing the region $R_{i+j}$, where 
 \begin{align}
 &R_{i+j} = \textit{Max}(R_{i}, R_{j}), \notag \\
    \textbf{if  }& \quad  \textit{IoU}(R_{i}, R_{j}) > 0.9 \notag\\ 
    & \And \textit{Parent-Child}(R_i, R_j) == 1. 
\end{align} 
Notably, we find that the merged connected nodes $R_{i+j}$ often indicate regions having united semantic contents due to their multiple code attributes. Therefore, we further label those merged nodes with more than one leaf as global regions and the corresponding leaves as local regions:
\begin{align}
 \text{Global Regions} = \notag \\ \{ R_{i+j}& | \textit{len}(\textit{Leaf}(R_{i+j})) > 1 \} \\
 \text{Local Regions} =  \notag \\ \{ \textit{Leaf}(R_{i+j})& | \textit{len}(\textit{Leaf}(R_{i+j})) > 1 \},
\end{align}
where  $\textit{Leaf}(R_i)$ outputs all the leaves in the tree tracing down from the node of region $R_i$.


\begin{figure}[t]
\includegraphics[width=\linewidth]{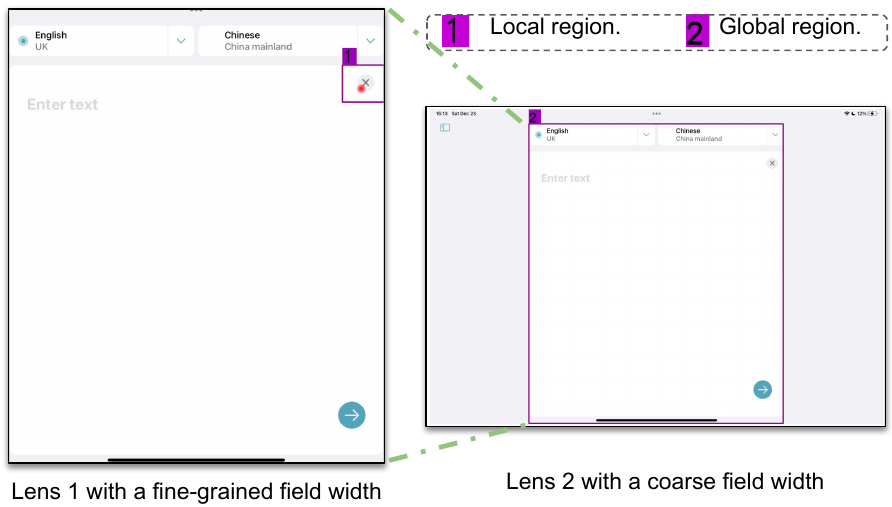}
\caption{Example of the lenses generated from the Hierarchical Layout Tree based on a point coordinate. Lens 2 can be seen as a zooming-out from Lens 1. ToL agent's corresponding output is in Appendix \ref{e}.}
\label{col_e}
\end{figure}

\subsection{Target Path Selection and Multi-lense Prompting} \label{chain-of-lens-prompting}
Based on the constructed Hierarchical Layout Tree, we next select a target path from the tree and generate a series of image prompts with annotations based on the input screenshot $S_i$ and point $P_i$. The image series represents different "lenses" with varying field widths and becomes the visual prompt for the MLLM, GPT-4o, which we refer to as Multi-lense prompting.

The process first identifies the smallest local region that contains the input point. Then the connected parent global region is extracted from the Hierarchical Layout Tree\footnote{In some rare cases, the input point is outside all local regions, we treat the global regions as local regions in this process and the parent of the global region is the full screenshot.}. We next curate two lenses in a sequence that visually articulates this hierarchical relationship from a fine-grained to a coarse view: the first lens shows only the global region from the screenshot while marking the local region with a colored bounding box labeled `1', the target point with a semi-transparent red dot; the second lens includes the complete input screenshot and marks the global region with another bounding box labeled `2'. Figure \ref{col_e} shows an example.

Then a Multi-lens Prompting method is adopted, where the curated set of lenses is sent to GPT-4o with a text prompt which includes the text coordinate of the target point. Some specific instructions are also included in the prompt, asking, "Explain where box 1 is in relation to box 2 and where box 2 is located on the overall screen." In this way, we provide clear visual cues that facilitate the GPT-4o to generate detailed descriptions that reflect the GUI's content and layout. For more details about the prompt used, please refer to Appendix \ref{visual_prompt_ToL agent}. 

\section{Screen Point-and-Read Benchmark}
\subsection{Overview}

\begin{figure}[t]
\setlength{\abovecaptionskip}{0.1cm}
\begin{subfigure}{.235\textwidth}
  \includegraphics[width=\linewidth]{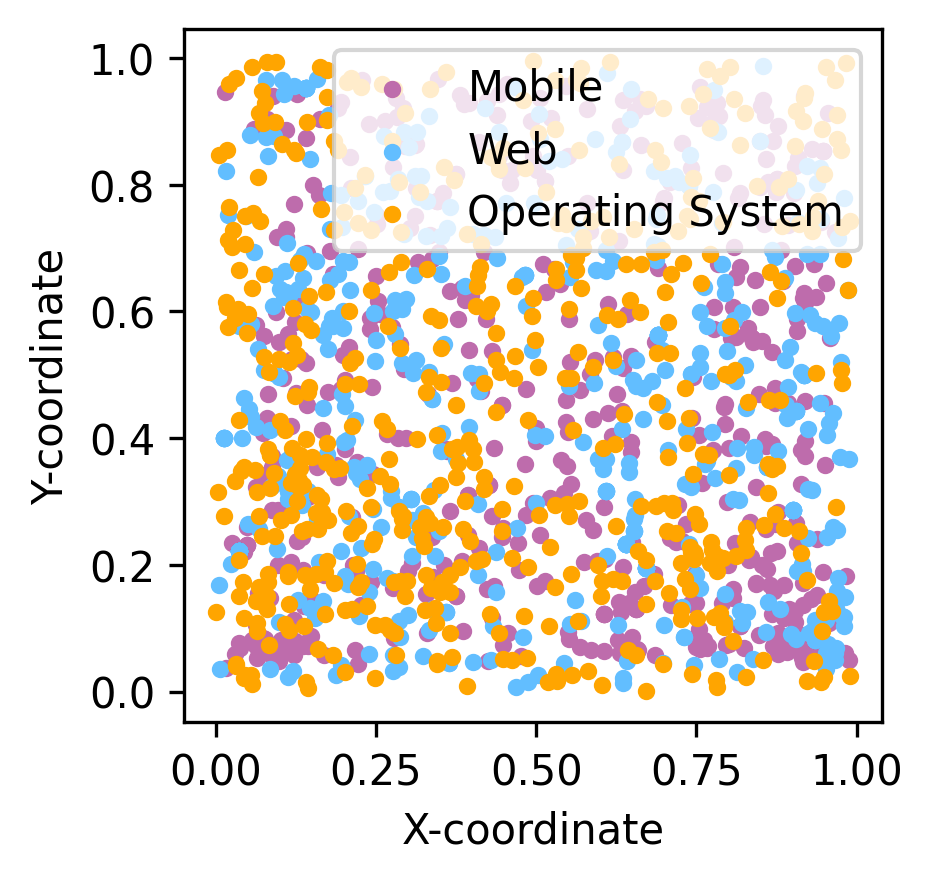}
  \caption{}
  \label{stat_1}
\end{subfigure} \hfill
\begin{subfigure}{.235\textwidth}
  \includegraphics[width=\linewidth]{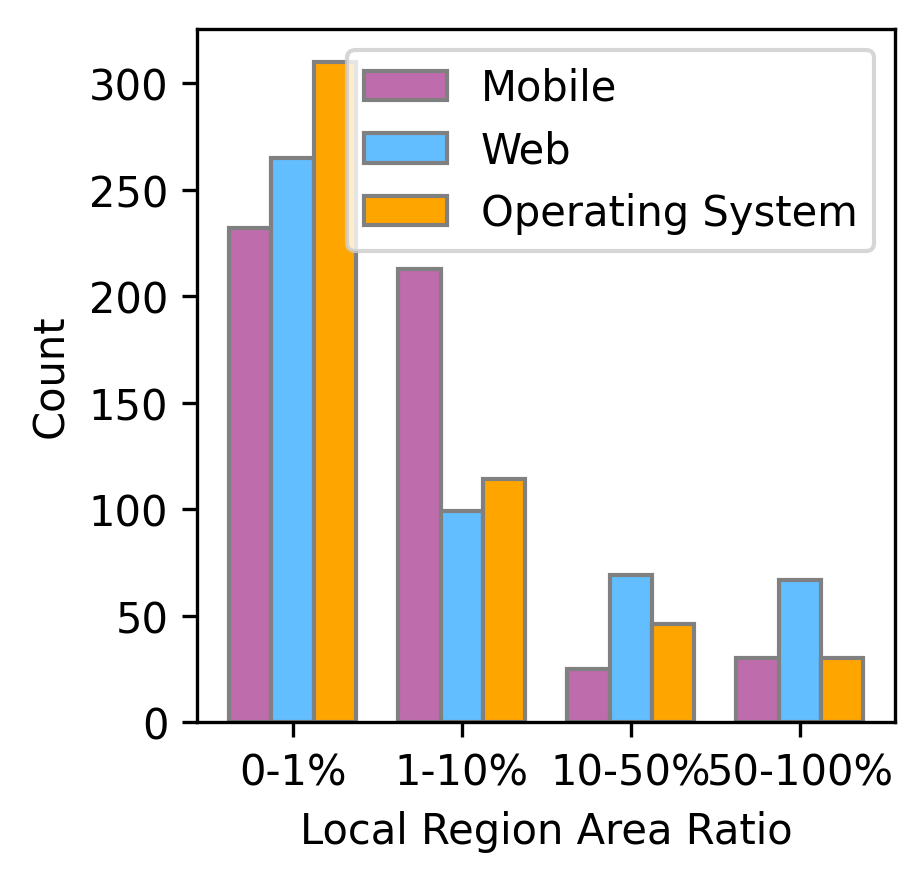}
  \caption{}
  \label{stat_2}
\end{subfigure}

\caption{(i) shows the normalized locations of target points $P_i$ in the input screenshot $S_i$. (ii) shows distributions of the area of local regions $R_i$ in our ScreenPR benchmark.}\label{fig:roc}
\end{figure}

We introduce the Screen Point-and-Read (ScreenPR) benchmark to rigorously evaluate our ToL agent on the ScreenPR task, where descriptions are required based on user-indicated points on screenshots. As shown in Table \ref{table:ScreenPR_stats}, this benchmark covers a diverse domain of GUIs with a total 650 screenshots. Each screenshot is accompanied by manually verified target points \( P_i \), their corresponding local regions \( R_i \), and content descriptions. Figure \ref{stat_1} illustrates the normalized coordinates of target points \( P_i \) in the ScreenPR benchmark, demonstrating comprehensive coverage of every location on the screen. The regions included have various content, and their area ranges from 1\% to over 50\% of the screen, as shown in Figure \ref{stat_2}.


\begin{table}[t]
  \setlength\tabcolsep{2pt}
  \centering
  \resizebox{\columnwidth}{!}{
\begin{tabular}{l|ccc}
\toprule
Domain & Screenshots & Target Points & Local Regions \\
\midrule
Web & 199 & 500 & 500 \\
Mobile & 201 & 500 & 500 \\
Operating System & 250 & 500 & 500 \\
\bottomrule
Total & 650 & 1,500 & 1,500 \\
\bottomrule
\end{tabular}
}
\caption{Key statistics of the ScreenPR Benchmark}
\label{table:ScreenPR_stats}
\end{table}

\subsection{Data Collection}
\label{data_collection}
We first gather mobile and web screenshots from Screenspot \cite{cheng2024seeclick} and operating system screenshots from the data explorer videos\footnote{\url{https://os-world.github.io/explorer.html}} of OSWorld \cite{OSWorld}. Then we randomly generate 4 or 5 points as the candidates for target points \( P_i \) within each screenshot. Next, we hire 3 students as annotators to draw bounding boxes for the local regions \( R_i \) corresponding to the points. To maximize the variety of the data, candidate target points that represent duplicated local regions will be removed. Finally, we assign the remaining points within each screenshot as either target points \( P_i \) or reference points \( P_{i'} \), where they are paired such that every pair of \( ( P_i, P_{i'} )\) all have non-overlapping corresponding local regions (\( P_{i'} \) is used for our proposed {\it cycle consistency evaluation} detailed in Section \ref{eval_method}). Additionally, we collect human-verified descriptions for all local regions \( R_i \) corresponding to \( P_i \). These descriptions are captions generated using GPT-4o with the direct input of $Crop(S_i,R_i)$, and we manually examined and corrected mistakes. 
After the annotation, we ensure the data quality by first using scripts to spot errors, such as ensuring that local regions \( R_i \) are included \( P_i \) inside, and then manually examine the data.

\subsection{Evaluation}
\label{eval_method}
The ScreenPR benchmark evaluates both the accuracy of the generated content description $\hat{D}^c$ and the generated layout description $\hat{D}^l$. Since there is no ground truth for the generated descriptions, we involve human evaluations and also propose automated cycle consistency evaluations inspired by the two-agent evaluation in \citet{padmakumar2022teach,fan-etal-2023-r2h}. 
\paragraph{Human evaluation} We conduct surveys with human judges to assess the quality of the generated descriptions. Each survey presents a screenshot with an indicated point, accompanied by descriptions from different screen readers. The judges are asked to rate these descriptions based on three criteria: how well they describe the content, how well they articulate the layout, and their overall preference. Each question has four options, ranging from "very well" to "not at all," which we numerically mapped to accuracy scores of 100\%, 66\%, 33\%, and 0\%, respectively.

\paragraph{Cycle consistency evaluation}
In order to evaluate screen reader models automatically, we propose the cycle consistency evaluation. The outputs $(\hat{D^c}, \hat{D^l})$ from the screen reader are each used as the input for an auxiliary model tasked with specific validation tasks, with the alternative model's performance indicating the quality of the screen reader's descriptions. 

Specifically, to evaluate content accuracy, GPT-4V \cite{achiam2023gpt} serves as the auxiliary model, performing a multi-choice selection task. This alternative model is asked to select one from the multi-choice candidates based on the predicted descriptions $\hat{D^c}$. The candidates are four screenshots of different GUI regions cropped based on human annotation, with only one targeted region 
\begin{align}
\text{Multi-choice}&\text{ question} = Q\\
    \text{Multi-choice}&\text{ candidates}(S_i, P_i, \hat{D^c}_i) \notag\\ = \{\textit{Crop}&(S_{k_{j}}, R_{k_{j}}) | j \in [0,4],   \notag\\
    &k_0 = i, k_1, k_2, k_3 \in N\}, 
\end{align}
where $Q$ is the fixed question and $\textit{Crop}(S_i,R_i)$ means the screenshot $S_i$ cropped at the region $R_i$.

For layout description accuracy, we design another multi-choice selection task for the auxiliary model, GPT-4 \cite{achiam2023gpt}. This multi-choice selection task requires the auxiliary model to select the positional relationship of two regions indicated by $P_i$ and $P_{i'}$ based on corresponding generated layout descriptions $\hat{D^l}_i$ and $\hat{D^l}_{i'}$.
$P_{i'}$, the reference points, collected in the same way as target points $P_i$ (detailed in Section \ref{data_collection}) are manually annotated to represent a region $R_{i'}$ that does not overlap with the local region of the target point $R_i$ on the same screen $S_i$, especially for the evaluation of the layout description accuracy.
The multi-choice selection task presented to the auxiliary model is

\begin{align}
&\text{Multi-choice question} = Q_\text{gen}(\hat{D^l}_i, \hat{D^l}_{i'}) \\
    &\text{Multi-choice candidates}  = C, 
\end{align}
where $Q_\text{gen}$ is a question template that needs to be filled with the layout descriptions generated, and $C$ are some fixed choices provided such as `upper,' `lower,' `left,' etc. Based on the ground truth location of $R_i$ and $R_{i'}$, we calculate the accuracy of the choices made by the auxiliary model, which indicates the screen reader's proficiency in describing the layout.



\begin{table*}[t]
\centering
\setlength\tabcolsep{3pt}
\resizebox{\textwidth}{!}{
\begin{tabular}{l|c|ccc|cc|cc} 
\toprule
           & \multirow{2}{*}{\textbf{Avg. Len. }} & \multicolumn{3}{c|}{\textbf{Human Evaluation~(\%)}}                                                                                         & \multicolumn{2}{c|}{\textbf{Cycle Consistency Evaluation~(\%)}}                                          & \multicolumn{2}{c}{\textbf{Language Similarity}}  \\ 
\cline{3-9}
           &                                      & \textbf{Content} & \textbf{Layout} & \textbf{Overall} & \textbf{Content Acc} & \textbf{Layout Acc} & \textbf{BERTScore} & \textbf{ROUGE}               \\ 
\hline
LlaVA-NeXT & 51.95                                & 21.69                                       & 9.59                                      & 21.19                                                & 25.43                                            & 7.20                                             & 85.18              & 16.52                        \\
CogAgent   & 30.69                                & 22.18                                       & 13.33                                      & 20.62                                                & 25.86                                            & 8.00                                             & 82.31              & 15.49                        \\
GPT-4o     & 34.44                                & 32.73                                       & 28.12                                      & 30.88                                                & 35.10                                            & 21.87                                            & 85.76              & 15.08                        \\
GPT-4o+Scaffolding & 28.51 & 35.21 &29.54 & 33.28 & 47.82 & 25.60 & 86.25 &17.25 \\
ToL Agent  & 38.37                                & \textbf{41.82}                                       & \textbf{33.42}                                      & \textbf{41.63}                                                & \textbf{74.96}                                   & \textbf{39.67}                                   & \textbf{86.73}     & \textbf{19.47}               \\
\bottomrule
\end{tabular}}

\caption{Main results. We evaluate our ToL agent against three other baselines with the human evaluation and cycle consistency evaluation of our ScreenPR benchmark. Additionally, we compare the generated content descriptions with human-verified content descriptions using language similarity scores. The results show that the ToL agent achieves the best performance. }\label{table:main}
\end{table*}

\begin{table}[t]
\setlength\tabcolsep{2pt}

\centering
\resizebox{0.5\textwidth}{!}{
\begin{tabular}{l|c|c} 
\toprule
 & \textbf{Content Acc} $$ & \textbf{Layout Acc} $$  \\ 
\hline
GPT-4o                                                                                            & 35.10                            & 21.87                                                    \\
ToL Agent (GPT-4o)                                                                     & \textbf{74.96}                     & \textbf{39.67}   \\                                          
\midrule
Qwen-VL-Plus                                                                      & 28.29                    & 12.50                                          \\
ToL Agent (Qwen-VL-Plus) & \textbf{41.93}                     & \textbf{14.47} \\
\midrule
Gemini Pro Vision  &33.20	&12.27\\
ToL Agent (Gemini Pro Vision)&	\textbf{45.33}	&\textbf{18.53}\\

\bottomrule
\end{tabular}
}
\quad

\caption{Results of the ToL Agent built upon various MLLMs. The results of our ToL agent based on various MLLMs are consistently better than the MLLMs alone, providing strong evidence that our approach is effective and generalizable. All the MLLMs are accessed through their APIs around Jun 2024.}
\label{table:generalizable}
\end{table}

\begin{table}[t]
\setlength\tabcolsep{2pt}

\centering
\resizebox{0.5\textwidth}{!}{
\begin{tabular}{l|c|c} 
\toprule
     & \begin{tabular}[c]{@{}c@{}}\textbf{Content}\\ \textbf{Acc} \end{tabular}  & \begin{tabular}[c]{@{}c@{}}\textbf{Layout}\\ \textbf{Acc} \end{tabular}                      \\ 
\hline
ToL Agent                                                                                            & 75.0                            & 39.7                                                    \\
\quad + w/o multi-lens visual prompt                                                                       & 71.2 ($-3.8$)                     & 37.7 ($-2.0$)                                             \\
\quad + w/o target point mark  ($\approx$ \text{SoM})                                                                      & 65.6~ ($-9.4$)                    & 37.6 ($-2.1$)                                             \\
 \begin{tabular}[c]{@{}c@{}}\quad+ w/o local \& global regions marks\\ (degrade to vanilla GPT-4o)  \end{tabular}   & 35.1 ($-29.9$)                       & 21.9 ($-18.6$) \\
\bottomrule
\end{tabular}
}
\quad

\caption{Ablation results. We gradually refine the ablation to exclude various processes in our ToL grounding mechanism for the ToL agent on the ScreenPR benchmark. The results show the effectiveness of various components.}
\label{table:ablation_0}
\end{table}

\section{Experiments}
\paragraph{Baselines}

We adopt several baselines, including generalist MLLMs: the proprietary GPT-4o \cite{gpt-4o} and the open-source LlaVA-NeXT \cite{liu2024llavanext}. We also test the CogAgent \cite{hong2023cogagent}, a MLLM specifically tuned for GUI tasks, and a visual prompting method Scaffolding \cite{lei2024scaffolding} that is specially developed to enhance the spacial reasoning ability of MLLMs. The inputs for baselines are the text prompt with the input point coordinate $P_i$ and the screenshot $S_i$. We keep prompts for baselines in the same style as the ToL agent, except that we have to simplify it for CogAgent as we find it necessary for CogAgent to follow the instructions. Detailed prompt design is shown in Appendix \ref{prompts_baselines_tol_agent}. 

\paragraph{Training}
We divide the ASHL dataset into 90 \% training and 10\% evaluation sets for the GUI region detection model training process.
Training is performed on four NVIDIA A6000 GPUs with a batch size of 8, spanning 90 epochs. The optimized model demonstrated excellent performance, achieving an Average Precision (AP) of 94.1\% and an Average Recall (AR) of 95.9\%.

\subsection{Main Results}
We evaluate our ToL agent against three baselines on the ScreenPR benchmark as shown in Table \ref{table:main}. The results are consistent across both human evaluation and automatic cycle consistency evaluation, showing that our ToL agent achieves the best performance in terms of content and layout description accuracy. Notably, our ToL agent demonstrates over a 15\% improvement compared to the GPT-4o model. Furthermore, the overall scores from the human evaluation indicate that descriptions generated by our ToL agent are the most favorable. Interestingly, although the results show that the CogAgent performs slightly worse than LlaVA-NeXT in terms of the content and layout description accuracy, the overall scores from human judges slightly prefer LlaVA-NeXT. We believe this preference is due to the LlaVA-NeXT model's tendency to produce longer descriptions with more information about the screen, creating a more comprehensive feel.

Additionally, we leverage language similarity scores, including BERTScore and ROUGE-L, to compare the generated content descriptions with human-verified descriptions. We observe that the overall trend from the language similarity evaluation results matches the content evaluation results from human and cycle consistency evaluations. However, the differences in language similarity scores between models are relatively indistinguishable. We believe this is because the captions are more comprehensive, sometimes redundant, whereas the screen reading output is more focused on delivering the key features of the content to help users understand it.

To further evaluate the generalizability of our ToL agent pipeline, we conduct experiments using the cycle consistency evaluation on our ToL agent with two other different MLLMs rather than the default GPT-4o: the Qwen-VL-Plus model \cite{bai2023qwen} and the Gemini Pro Vision model \cite{team2023gemini}. We access both of these models via their APIs. As shown in Table \ref{table:generalizable}, the results show that our ToL agent built upon various MLLMs consistently outperforms the standalone MLLMs, offering compelling evidence that our approach is effective and generalizable.

\begin{table}[t]
\setlength\tabcolsep{1pt}
\centering
\resizebox{0.5\textwidth}{!}{
\begin{tabular}{l|c|c} 
\toprule
                         & \textbf{Content Acc} $$ & \textbf{Layout Acc} $$  \\ 
\hline
Human                    & 92.8                             & 70.3                            \\ 

ToL Agent  & 72.5 (-20.3)                            & 41.3  (-29.0)                          \\
ToL w/ GT local regions  & 85.5 (-7.3)                           & 44.3 (-26.0)                           \\
ToL  w/ GT global regions & 88.2 (-4.6)                           & 46.6  (-23.7)                          \\
ToL w/ GT global  local regions & 89.5 (-3.3)                           & 48.2 (-22.1)                           \\
\bottomrule
\end{tabular}
}

\caption{Bottleneck analysis. Replacing local and global regions with GT annotations mainly improves content description accuracy, indicating rooms of improvements for both the GUI region generation process and the MLLM leveraged.}
\label{table:ablation_1}
\end{table}

\begin{figure*}[t]

    \includegraphics[width = \textwidth]{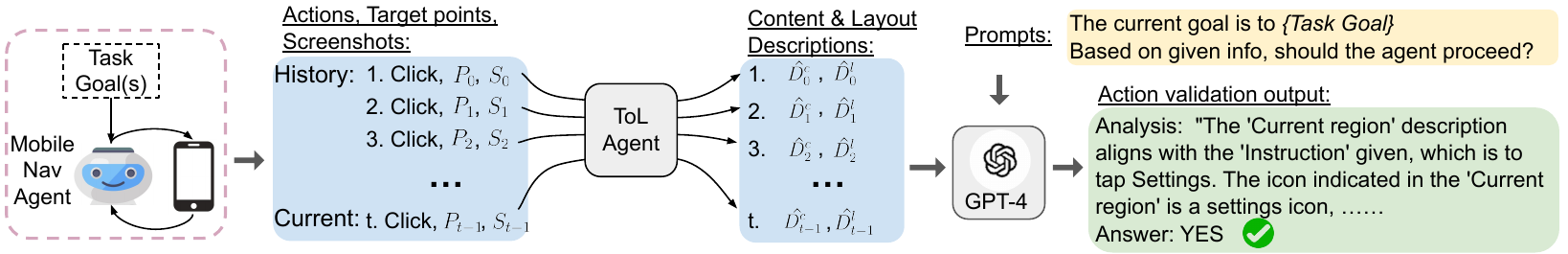}
    \caption{Pipeline of employing our ToL agent in verifying the actions from a mobile navigation agent.}
    \label{fig:application}
\end{figure*}

\subsection{Ablation Study}

\paragraph{Effectiveness analysis} To assess the effectiveness of the ToL design, we conduct an ablation study where we progressively refine the ablation to exclude key design components and evaluate these ablated versions on our ScreenPR benchmark using the cycle consistency evaluation, as shown in Table \ref{table:ablation_0}.
First, we modify our ToL agent to exclude the multi-lens visual prompting. This change necessitates annotating bounding boxes for the local region, global region, and the target point on the complete screenshot as a single visual prompt. Then we further remove the red dot mark from the visual prompt, resulting in a design similar to the Set-of-Mark (SoM) prompting method \cite{yang2023setofmark}. As a result, these changes lead in performance degradation in both content and layout description accuracy, indicating the effectiveness of our visual prompting strategies.
Finally, we ablate the input visual prompt to remove marks of local and global regions that we extract from our Hierarchical Layout Tree. This drastic modification results in a significant decrease in performance, underscoring the importance of our Hierarchical Layout Tree in the of the ToL design.

\paragraph{Bottleneck analysis} To examine the bottlenecks in our ToL agent, we replace the local and global regions generated from the Hierarchical Layout Tree with human annotations, considered as the ground truth (GT). We also assess human performance by having human annotators write both content and layout descriptions manually. Due to the high cost of human efforts, we conduct experiments on a 20\% random sample of the ScreenPR benchmark.
As shown in Table \ref{table:ablation_1}, substituting either GT local regions or GT global regions significantly close the gap toward human performance in the ToL agent's accuracy of content descriptions. However, for layout descriptions, the gap remains, indicating that the MLLM we leverage is the bottleneck in this aspect. This finding aligns with the conclusion from \cite{fan2024muffin}, which highlights that current MLLMs face significant challenges in understanding the layout of multipanel images.

\subsection{Broader Application: Verification of Mobile Navigation Agent Actions}

In this section, we demonstrate how our ToL agent helps identify incorrect actions for a mobile navigation agent, MagicWonder, on the MagicWand platform~\cite{dingenhancing}. Driven by a specific task goal, such as "connect to Wifi in settings", MagicWonder can generate actions to interact with a live mobile screen and save the detailed information for each step along its execution trajectory. More details are provided in Appendix \ref{action_v}. To incorporate our ToL agent for verification of MagicWonder's output actions, we design a simple pipeline. As shown in Figure \ref{fig:application}, the layout and content descriptions are generated by our ToL agent for each action-point-screenshot triplet and sent to the GPT-4 model. The GPT-4 model, based on the task goal, instructions, and descriptions from the ToL agent, decides whether the last action from the MagicWonder is correct.

\paragraph{Data sampling} We select 52 mobile agent trajectories with specific task goals from instruction-guided executions on the MagicWand platform, with more details in Appendix \ref{execution_trajectory}. Each trajectory consists of multiple execution steps triggered by distinct actions, resulting in a total action number of 209. We hire two students to annotate incorrect actions in the chosen trajectories to get the ground truth labels.

\paragraph{Result} We perform 10 end-to-end evaluations and present the mean value and standard deviation in Table \ref{tab:action_validation}. Despite minor fluctuations, while 92\% correct actions can be stably recognized, about 38\% incorrect actions can be identified. This reveals that by incorporating the ToL agent, the performance of the mobile navigation agent is significantly and robustly improved.
Our ToL agent can also identify particularly tricky challenges faced by mobile navigation agents during execution. One notable issue is the "execution loop", where agents repeatedly take actions on the same or similar regions without making progress. Our results show that for all 20 incorrect actions that fall into this category, 44\% ± 0.04 of them can be successfully detected. To further validate the effectiveness of our method, we introduce two baselines: Confidence Score Filter and GPT-4o Verifier. The Confidence Score Filter considers actions as positive when the relevant confidence score is above 0.7 and the GPT-4o Verifier directly sends the actions with corresponding screenshots as prompts to GPT-4o to identify the correctness of actions. We run the end-to-end evaluation equivalently and the results in Table \ref{tab:action_verification_of_baselines} show that our method of using the ToL agent to verify the output from the mobile navigation agent outperforms the other baselines.

\begin{table}[t]
    \centering

    \setlength\tabcolsep{4pt}
    \resizebox{\columnwidth}{!}{

\begin{tabular}{c|ccc} 
\toprule
                   & \multicolumn{2}{c|}{\textbf{Predicted as}}                     & \multirow{2}{*}{\textbf{TP/FP Rate}}  \\ 
\cline{2-3}
                   & $\textbf{Correct}$ & \multicolumn{1}{c|}{$\textbf{Incorrect}$} &                                       \\ 
\cline{2-4}
$\textbf{Correct Action}$   & 100.2$\pm$2.0      & 8.8~$\pm$2.0                              & 92\% $\pm$0.02                        \\
$\textbf{Incorrect Action}$ & 61.9~$\pm$2.9      & 36.1~$\pm$2.9                             & 38\% $\pm$0.03                        \\
\bottomrule
\end{tabular}
    }
  \caption{Results of leveraging the ToL agent to verify actions from a mobile navigation agent. "Predicted as" column shows the number of predicted actions. True positive and false positive rate are shown on the right most column calculated from the data on the left} 
  \label{tab:action_validation}
\end{table}

\begin{table}[t]
    \centering
    \setlength\tabcolsep{4pt}
    \resizebox{\columnwidth}{!}{
    \begin{tabular}{l|c|c} 
    \toprule
    \textbf{Verification Method} & \textbf{F1 score} & \textbf{Repetition Detection Rate} \\
    \hline
    Confidence Score Filter & 65\%$\pm$0.00 & 0\%$\pm$0.00 \\
    GPT-4o                           & 61\%$\pm$0.01 & 40\%$\pm$0.10 \\
    ToL Agent                             & \textbf{74\%$\pm$0.01} & \textbf{44\%$\pm$0.04} \\
    \bottomrule
    \end{tabular}
    }
  \caption{Performance of different verification methods in verifying actions of mobile navigation agent. The result shows that using our ToL agent to verify the actions of the mobile navigation agent achieves the best F1 score and performs the best in detecting the notable issue of the "execution loop", i.e. repetition detection.} 
  \label{tab:action_verification_of_baselines}
\end{table}

\section{Conclusion}

In this work, we propose a novel Tree-of-Lens (ToL) grounding method and build a ToL agent for the Screen Point-and-Read (ScreenPR) task, an important GUI referring task that is critical for accessible technology. Our ToL agent outputs layout-aware descriptions for pointed regions on screenshots.
With our newly proposed Screen Point-and-Read (ScreenPR) benchmark, we show that our ToL agent outperforms all other baselines in terms of content and layout description accuracy, demonstrating great potential in assisting visually impaired users when using digital devices.
Last but not least, in our experiment, we also apply our ToL agent to identify incorrect actions performed by a mobile navigation agent, yielding promising results.

\section{Limitations}

In this work, we propose a novel Tree-of-Lens (ToL) agent for the Screen Point-and-Read (ScreenPR) task. By leveraging GPT4-o to generate content and layout descriptions, we are able to set up the pipeline quickly and validate the effectiveness of ToL grounding method in a short time period. However, the accumulated server delay in requesting GPT4-o services has occupied more than 80\% of end-to-end evaluation, not to mention the high expense accompanying, which is a non-trivial issue for real-world applications. Thus, a more efficient and locally hosted model has to be given out in place of GPT-4o. Moreover, since our ToL agent relies on a certain MLLM to generate descriptions, there are potential risks that the generated descriptions from the MLLM could contain harmful contents. Therefore more post-processing strategy might be needed. Last but not least, another critical challenge is balancing application efficiency with improved semantic representation, thereby enabling on-device support and further expanding ToL's application areas. Our current work is limited for research purpose and more application-wise exploration is pending.

\bibliography{custom}

\appendix

\label{sec:appendix}
\newpage

\section{Example outputs of the ToL Agent}
\label{e}
\begin{figure*}
    \includegraphics[width=\textwidth]{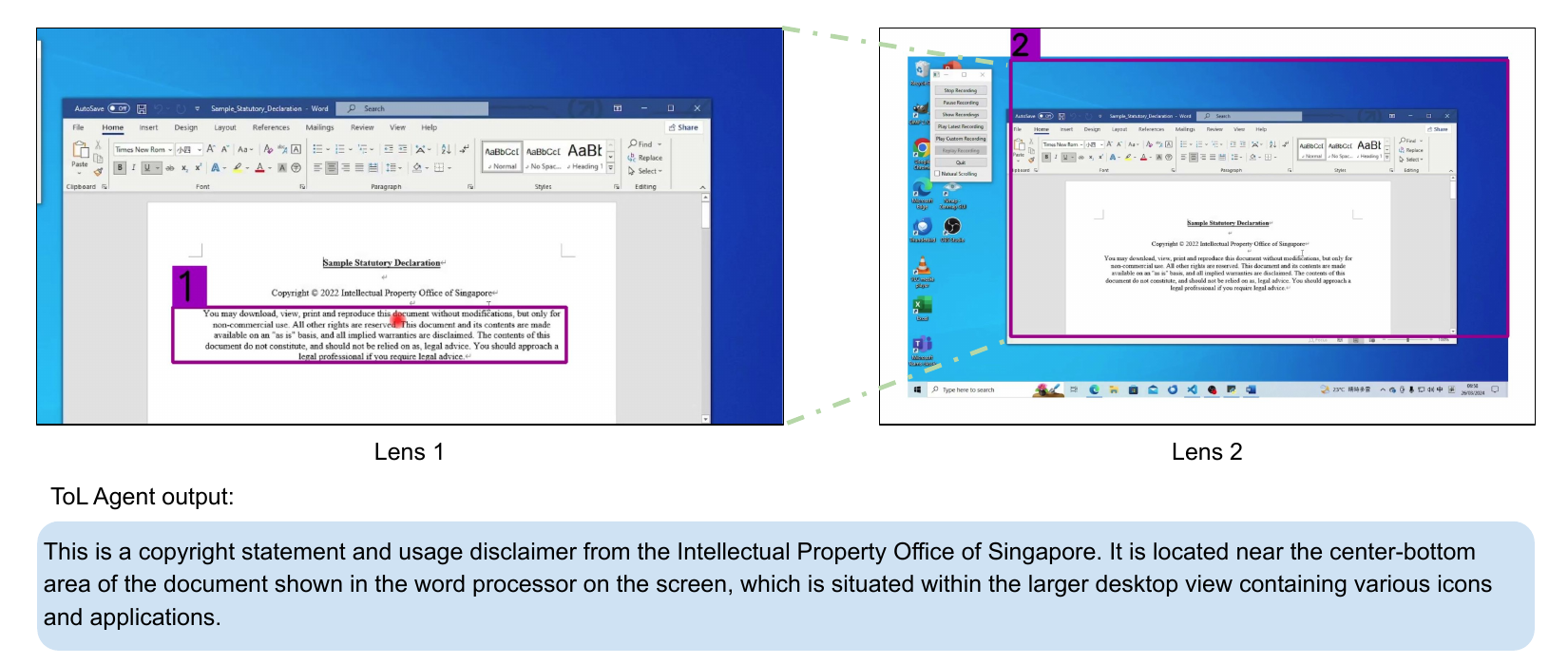}
    \caption{Example of the generated description for an operation system screenshot from our ToL agent with the two lenses generated. The global region generated is not perfect as the bounding box is a little bigger than the application window on the top right part. }
    \label{e1}
\end{figure*}

\begin{figure*}
    \includegraphics[width=\textwidth]{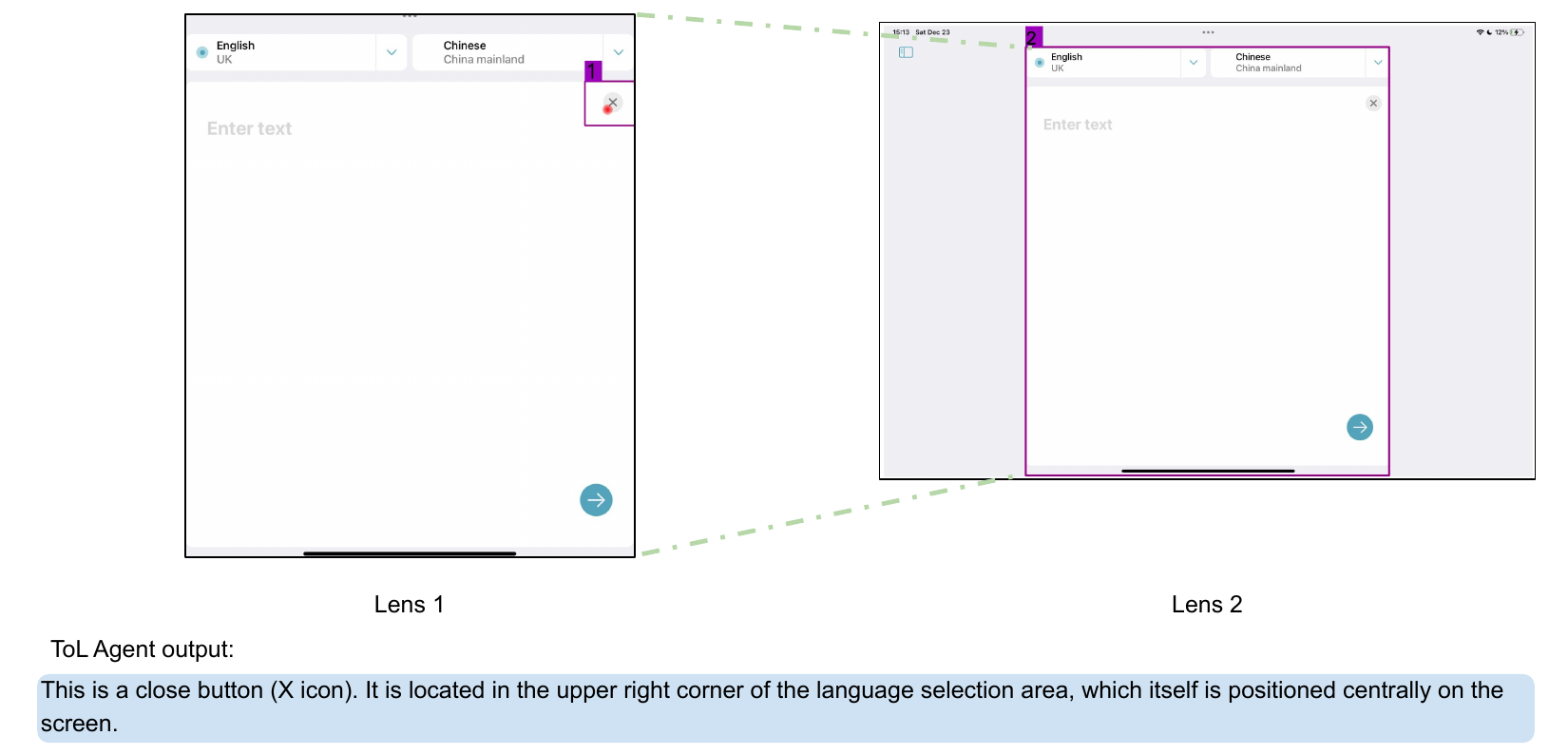}
    \caption{Example of the generated description for a mobile screenshot from our ToL agent with the two lenses generated.}
    \label{e2}
\end{figure*}

\begin{figure*}
    \includegraphics[width=\textwidth]{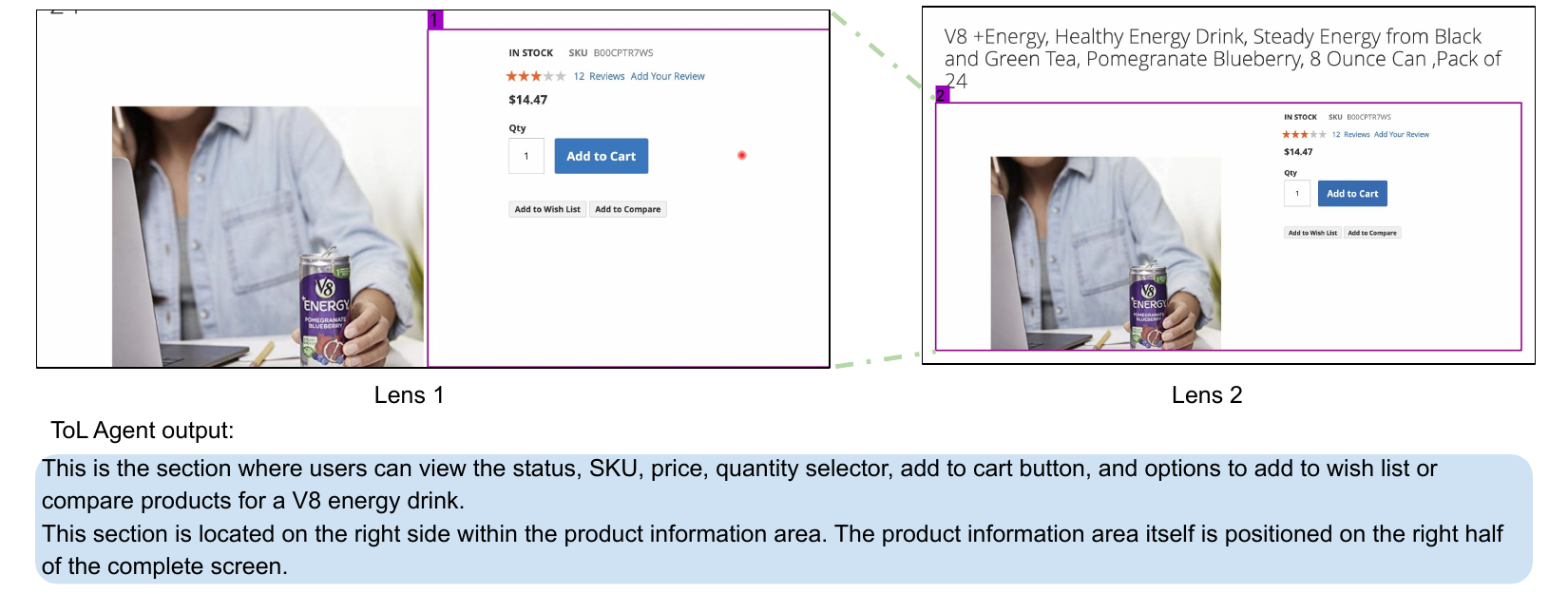}
    \caption{Example of the generated description for a web screenshot from our ToL agent with the two lenses generated. The description output is not perfect regarding the layout information as the "product information area" indicated by the global region is not only "on the right half of the complete screen" but the middle and lower areas of the complete screen. }
    \label{e3}
\end{figure*}

For each GUI domain, we show one example of the output of our ToL agent along with the lenses generated. We do not deliberately cherry-pick the best case but show examples with both strong and weak performances. An example for the operation system domain is shown in Figure \ref{e1}. An example for the mobile domain is shown in Figure \ref{e2}. An example for the web domain is shown in Figure \ref{e3}.


\section{Examples of the Cycle Consistency Evaluation}
\label{cce_example}
\begin{figure*}[t]
    \centering
    \includegraphics[width = 0.85\textwidth]{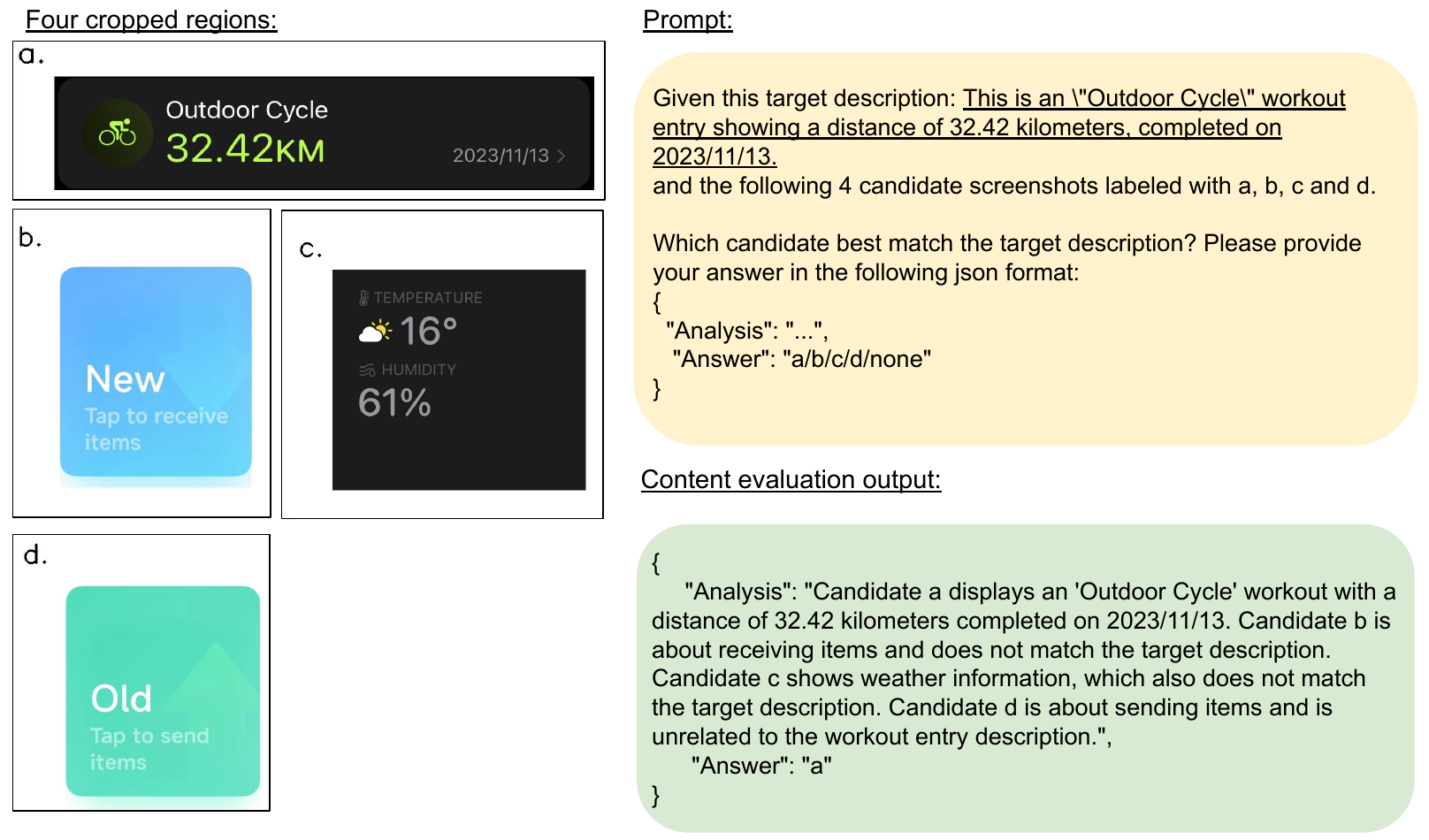}
\caption{One example of our cycle consistency evaluation for content description. A predicted description is shown on the right and four different cropped screen regions were sent as four separate images to GPT-4o. GPT-4o choose the matched region or answer with "unknown".}
\label{demo:content_evaluation}
\end{figure*}

We select one result of our cycle consistency evaluation for the content description and show it in Figure \ref{demo:content_evaluation}. The left four images come from the cropped areas from different GUI screens, with only one of which matches the description in the prompt on the right side. As shown, GPT-4o can successfully choose the correct option in the "Answer" field and give its reasoning in the "Analysis" field.

\begin{figure}[t]
    \centering
    \includegraphics[width = \columnwidth]{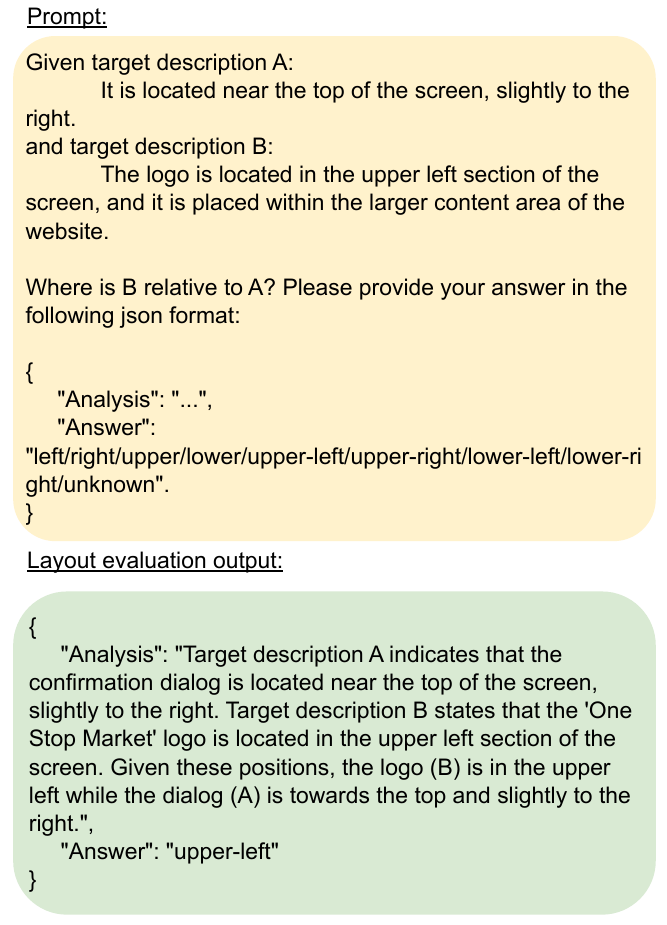}
\caption{One example of our cycle consistency evaluation for layout descriptions. The layout descriptions for target point and reference point are provided in prompt and GPT-4o is asked to decide their spatial relationship by choosing one option from the given 9 options.}
\label{demo:layout_evaluation}
\end{figure}

For cycle consistency evaluation regarding the layout description, the question contains the generated layout descriptions from models. We let GPT-4o choose one option from predefined 9 relative positions together with its analysis, referring to the layout evaluation output in Figure \ref{demo:layout_evaluation}. The ground truth is provided with script based on the human annotated local regions for the target point and reference point. The ground truth is also checked by human. 

\begin{figure}[th]
    \centering
    \includegraphics[width=\columnwidth]{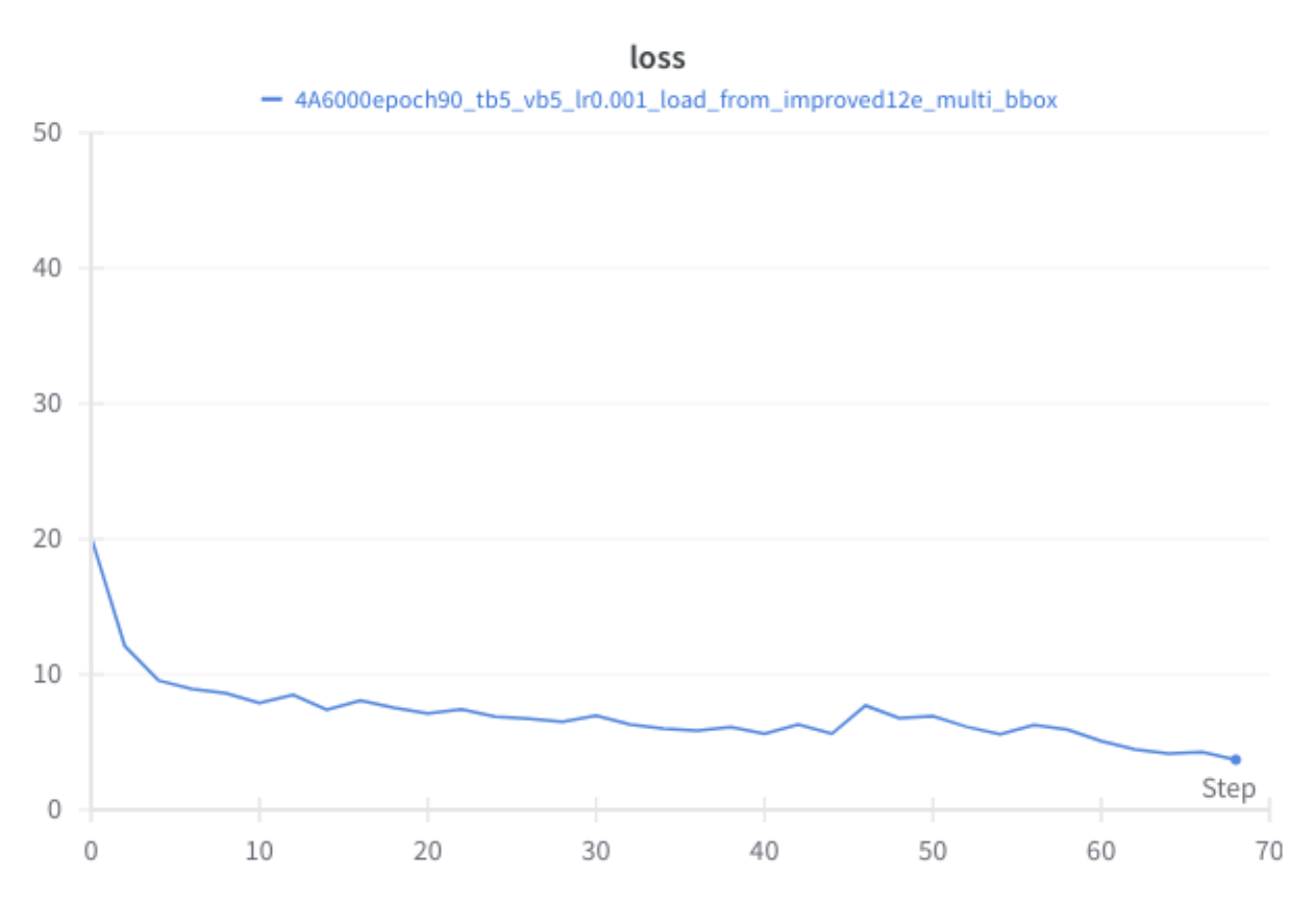}
    \caption{Training Loss of ResNet-50 backbone with epoch 90.}
    \label{fig:4}
\end{figure}

\begin{figure}[h] 
\begin{smallermdframed} 
You are a smart screen reader that outputs concise natural language to answer questions from users based on the area (box 1) pointed out by the user shown as a red dot on the screen. The red dot is inside the box 1 in the first image, at (x,y) = ($\{\{x\}\}$,$\{\{y\}\}$), where x and y are the normalized coordinates.  \par \text{\space} \par
Note: box 1 is the box with label 1 and box 2 is the box with label 2, box 1 is located inside box 2  \par
Note: the first image shows the box 1 from the view of box 2, and the second image shows the box 2 from the complete screen.  \par
Note: if the user asks about the location, based on the layout, explain where box 1 is in box 2 and then explain where box 2 is in the overall screen.   \par
Note: don't mention box 1, box 2 or the red dot in the output. \par \text{\space} \par
User question: (1) what is this? (2) where it is located in the screen? \par
Your output should in format (1) … (2) …
\end{smallermdframed}
\caption{Prompt used by ToL agent}
\label{prompt:ToL agent}
\end{figure}

\paragraph{Prompt used by LlaVA-NEXT and GPT-4o}

For the baseline models, LlaVA-NEXT and GPT-4o share the same prompting template as shown in Figure \ref{prompt:llava_next_gpt4o}. 

\begin{figure}[h] 
\begin{smallermdframed} 
You are a smart screen reader that outputs concise natural language to answer questions from users based on the area pointed by the user at (x,y) = ($\{\{x\}\}$,$\{\{y\}\}$), where x and y are the normalized coordinates of the image. \par \text{\space} \par
Note: if the user asks about the location, based on the layout, explain where the target area is located locally and then explain where it is in the overall screen.  \par
Note: don't mention the area directly in the output, instead, use "it", "this". \par \text{\space} \par
            
User question: (1) what is this? (2) where it is located in the screen?\par
Your output should in format (1) … (2) …\par
\end{smallermdframed}
\caption{Prompt used by LlaVA-NEXT and GPT-4o}
\label{prompt:llava_next_gpt4o}
\end{figure}

\paragraph{Prompt used by CogAgent}

In order to drive CogAgent to follow instructions, a simplified prompting template is applied, as shown in Figure \ref{prompt:cogagent}. 

\begin{figure}[ht] 
\begin{smallermdframed} 
What is indicated by the point at (x,y) = ($\{\{x\}\}$,$\{\{y\}\}$), where x and y are the normalized coordinates of the image. \par 
and where it is located in the screen? 
\end{smallermdframed}
\caption{Prompt used by CogAgent}
\label{prompt:cogagent}
\end{figure}

\begin{figure}[htbp] 
\begin{smallermdframed} 
Given the following information in a mobile navigation task:\par
Historical action and region description: \{\{haction\}\}\par
Task Goal: \{\{goal\}\}\par
Current region: \{\{region\}\} \par \text{\space} \par

The agent now is going to interact with the "Current region" with the action: \{\{action\}\}. Should the agent proceed? \par
Note: The agent should not proceed if the "Current region" is repeated too often in "Historical action and region description". \par
Note: The agent may proceed if the "Current region" aligns with "Task Goal". \par \text{\space} \par

Please provide your answer in the following JSON format:\par
\{\par
    "Analysis": "...",\par
    "Answer": "yes/no"\par
\}\par
\end{smallermdframed}
\caption{Prompt used by GPT-4o in Action Validation}
\end{figure}

\begin{figure*}[t]
    \centering
    \includegraphics[width = 0.85\textwidth]{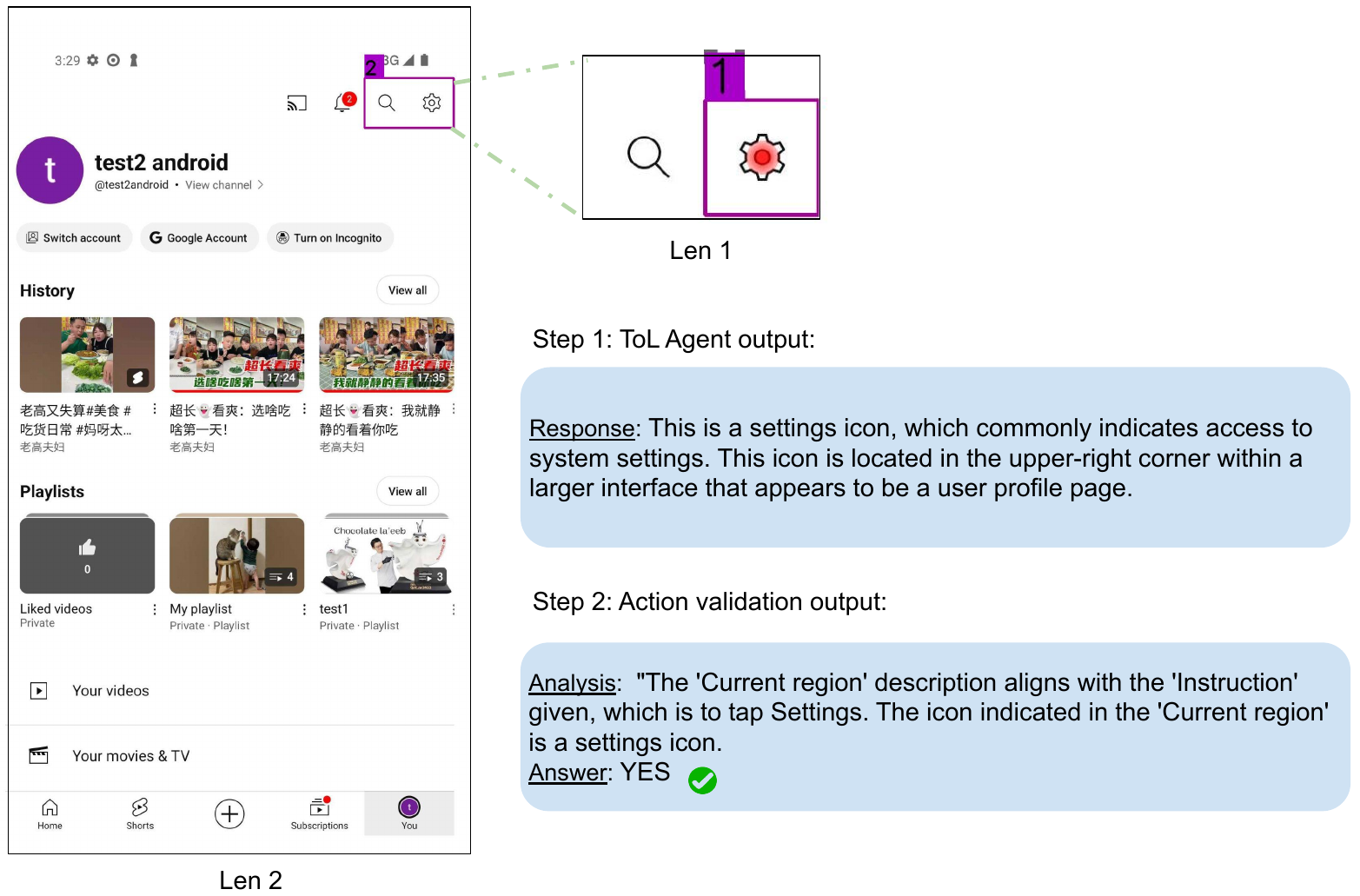}
\caption{Given the goal of "pausing historial track in Youtube App", mobile Agent on MagicWand plans to click on the screen. ToL Agent generates descriptions for the action planed. The description is sent as input for action verification reasoning, where GPT-4o model output the analysis with decision, which is shown on the bottom right. The GPT-4o concludes that the click action is correct.}
\label{demo:action_validation}
    
\end{figure*}

\section{Cross-evaluation Experiments with Cycle Consistency Evaluation}

Our cycle consistency evaluation, relying on GPT-4V as the auxiliary model, shows that our ToL agent (leveraging GPT-4o by default) outperforms the other baselines including the vanilla GPT-4o. Since the evaluator, the baseline model, and our ToL agent all rely on GPT models, we believe the results regarding models relying on GPT models are less influenced by the potential bias as mentioned by the review. Furthermore, to strengthen our result, we first include the Gemini Pro Vision model as a baseline model and introduce a new version of our ToL agent relying on the Gemini Pro Vision model. Then, we extend our main results with the cross-evaluation setting where the models relying on GPT-4o will not be evaluated with GPT-4V as the auxiliary model, and the models relying on Gemini Pro Vision will not be evaluated with Gemini Pro Vision as the auxiliary model. The results are shown in Table~\ref{tab:cross_validate}, where under this cross-evaluation setting, our ToL agent consistently demonstrates clear superiority over the vanilla MLLM baseline.

\begin{table*}[t]
  \setlength\tabcolsep{1pt}
  \centering
  \begin{minipage}{0.5\textwidth}
    \centering
    \resizebox{\textwidth}{!}{
      \begin{tabular}{l|c|c}
\toprule
& \multicolumn{2}{c}{$\begin{array}{l}
    \text{Auxiliary Model: }  \\
     \text{GPT-4V}
\end{array}$} \\
\cline{2-3}
                         & \textbf{Content Acc}  & \textbf{Layout Acc}  \\ 
\hline

Gemini Pro Vision & 33.20	& 12.27                         \\
ToL Agent (Gemini Pro Vision) &	\textbf{45.33}	& \textbf{18.53 }                       \\
\bottomrule
      \end{tabular}
    }
    
  \end{minipage}%
  \hfill
  \begin{minipage}{0.5\textwidth}
    \centering
    \resizebox{\textwidth}{!}{
      \begin{tabular}{l|c|c}
\toprule
& \multicolumn{2}{c}{$\begin{array}{l}
    \text{Auxiliary Model: }  \\
     \text{Gemini Pro Vision}
\end{array}$} \\
\cline{2-3}
                         & \textbf{Content Acc}  & \textbf{Layout Acc}   \\ 
\hline

GPT-4o & 43.52 &	22.33                       \\
ToL Agent (GPT-4o) $\quad \quad \quad \quad \quad$ &	\textbf{58.44} &	\textbf{33.93}
         \\
\bottomrule
      \end{tabular}
    }
  \end{minipage}
  \caption{Cycle consistency evaluation of the ToL agent and vanilla MLLMs under the cross-evaluation settings. The performance of the ToL agent and the baseline MLLM are evaluated with the auxiliary model being a different MLLM than the one under test. The ToL agent consistently outperforms the baselines.}
  \label{tab:cross_validate}
\end{table*}

\section{Text Prompt templates} \label{prompts_baselines_tol_agent}

We show the templates of text prompts we used for different MLLMs. We substitute the $\{\{x\}\}$,$\{\{y\}\}$ with the normalized coordinate of the input point.

\paragraph{Prompt used by ToL agent} \label{visual_prompt_ToL agent} 

For our ToL agent, we use the text prompt template shown in Figure \ref{prompt:ToL agent} for GPT-4o.

\section{Mobile Navigation Action Verification}
\subsection{Example}
\label{action_v}
Figure \ref{demo:action_validation} illustrates one action verification case in our experiment: MagicWonder is asked to accomplish "pause historical track in Youtube App". Len 1 and Len 2 show the agent was attempting to click the "Setting" icon by giving the instruction "Tap Settings" at the second step along the execution trajectory. The first step in action verification is to get the response from the ToL Agent that describes Len 1 and Len 2 correctly with more detailed application contexts. Using this information, GPT-4o could correctly judge whether this action on the screen was in accordance with the given instruction.

\subsection{Sample Strategy of Mobile Navigation Action Execution Trajectory} \label{execution_trajectory}

MagicWand simulator can launch mobile agents to take action on devices under the guidance of a series of instructions. The corresponding runtime GUI context, like GUI screenshots, actionable regions, instructions for each step, and action details, are saved into a self-descriptive JSON format. We deliberately choose the instructions from the WeWeb dataset \cite{dingenhancing} having few overlapping with our ASHL dataset. After execution on MagicWand, we chose those trajectories with task completion rates less than 50\% to evaluate whether ToL information was accurate enough. 
The selected trajectories cover 11 Android applications from 8 application domains, where the chosen applications include SYSTEM, Entertainment, Education, Food \& Drinks, Communication, Shopping, News, Maps \& Navigation. The chosen Android apps include Messenger, Google Maps, YouTube, Settings, Quizlet, FlipBoard, McDonalds, eBay, Google Chat, Here WeGo, and DoorDash.

\subsection{Action verification Pipeline Details} \label{action_validation}

As shown in Figure \ref{fig:application}, a set of navigation actions, target points, and corresponding screenshots, are first recorded during the interaction between the mobile navigation agent and the mobile screen. Then, our ToL agent first processes each pair of target points and screenshots $(P_i,S_i)$, including those from the history path of navigation trajectory and the currently planned one  $(i\in[0, t-1])$, where the $P_t-1$ is the target point that the navigation agent plans to interact with for the next step. As a result, the layout and content descriptions $\hat{D}^c_i$, $\hat{D}^l_i$ are generated. Next, we concatenate all the descriptions and send them to the GPT-4 to determine whether agent actions follow the given task goal. 

\paragraph{Target path selection} For each mobile screenshot saved in the execution trajectory, the DINO detection model extracts all global regions with confidential scoring larger than 0.15 and local regions with confidential scoring larger than 0.05, consistent with our baseline evaluation. For click action, ToL first finds the smallest local and global regions that contain the click point. Usually, DINO's output aligns with regions that have clear pixel boundaries. To improve the accuracy of local regions with smaller sizes, we extend DINO's output regions by 50 pixels. If no match is found, we further extend the boundaries horizontally, following general GUI design principles. Different from click action, input action involves a region with a clear-cut boundary. Accordingly, ToL selects the smallest local region with an IoU value greater than 0.4 with the input region and the corresponding global region with an IoU value greater than 0.1 with the input region. With the chosen local and global regions, ToL follows Chain-of-Lens prompting mentioned in section \ref{chain-of-lens-prompting} to get region description.

\paragraph{Action verification} We prompt GPT-4o by using the current action name, instruction, and region description at the current step, together with historical action names and region descriptions, letting it judge whether the mobile agent can proceed. Only two general rules are mentioned in the prompt:

\begin{enumerate}
    \item The agent should not proceed if the "Current region" is repeated too often in "Historical action and region description"
    \item The agent may proceed if the "Current region" aligns with "Instruction"
\end{enumerate}

GPT4o's response includes a simple yes or no and a detailed reason.

\paragraph{Prompt used by GPT-4o in Action Verification} \label{prompt_AVAgent} 

To evaluate the role of LoT in mobile agents, we only list historical action and region description, current region, and the instruction for the current execution step together with the format we expect. During runtime, the aforementioned information will be used to replace variables in the prompting template \{\{haction\}\},  \{\{instruction\}\},  \{\{region\}\},  \{\{action\}\} separately.

We briefly mention two high-level goals that the agent is trying to achieve and expect it to figure out details via ToL information. More fine-grained work could be introduced later to get even better performance of action validation, though not our main focus now.

\section{Human Evaluation}

\begin{figure*}
    \includegraphics[width=\textwidth]{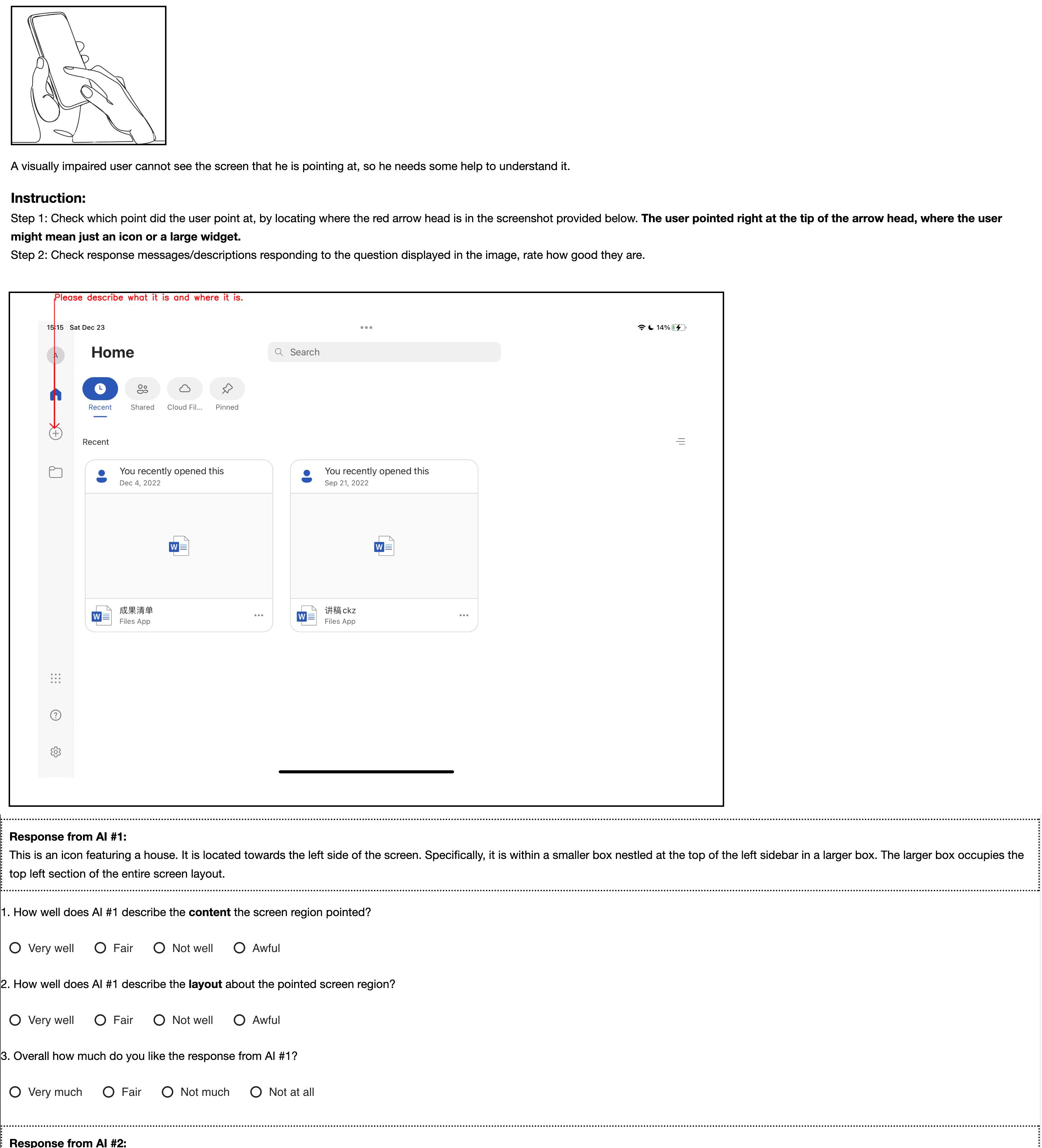}
    \caption{Screenshot of the survey we used in the human evaluation based on our ScreenPR benchmark. Due to the limited page length, questions are not shown completely, but the rest of the questions follow the same format as the one shown. }
    \label{append_survey}
\end{figure*}

We leverage human evaluation to test the performance of different screen readers on our ScreenPR benchmark. We leverage the Amazon Mechanical Turk\footnote{\url{https://www.mturk.com}} platform to find annotators to complete the survey that we designed for our human evaluation. We hired 3 human judges for each data sample. The judges are asked to rate the generated descriptions based on three criteria: how well they describe the content, how well they articulate the layout, and their overall preference. Each question has four options, ranging from "very well", “fair”, “not well”, and "awful," which we numerically mapped to accuracy scores of 100\%, 66\%, 33\%, and 0\%, respectively. We pay \$0.15 for each survey and the screenshot of the survey is shown in Figure
\ref{append_survey}. Approximately 81\% of the data samples receive at least two identical ratings from the human judges, and the standard deviation is 19.4\%.

\section{GUI Region Detection Model} \label{appendix:dino_traing_validation}

\paragraph{Training details} Using MMDetection framework \cite{MMDetection}, we change the training into two-label classification according to our task setting. Considering our dataset volume, we load pre-trained weights yielding the best performance on COCO, instead of training from scratch. 
We train the model for 90 epochs with a MultiStepLR scheduler set by milestones per 30 epochs and $\gamma$ = 0.1, the loss had further reduced to 2.5803 with small fluctuation, as shown in Figure \ref{fig:4}.

We also examine the Average Precision metrics with different IoU and Area on the validation set: they show a clear convergence trend after epoch 70, and the results on the validation set are shown in Table \ref{tab:mAPcoco}.

\begin{table}[t]
    \centering
    \resizebox{0.6\columnwidth}{!}{
    \begin{tabular}{cc|c}
        \toprule
        IoU       & Area   & Average \\ 
        \midrule
        0.50:0.95 & all    & 0.9410  \\
        0.50      & all    & 0.9620  \\
        0.75      & all    & 0.9470  \\
        0.50:0.95 & small  & 0.702   \\
        0.50:0.95 & medium & 0.897   \\
        0.50:0.95 & large  & 0.943   \\ 
        \bottomrule
    \end{tabular}}
    \caption{Average precision on the validation set for GUI region detection model. According to the definition in the Microsoft COCO dataset, the small and medium areas in an image are set as less than 1024 (32*32), 9216 (96*96), and others are large areas in the MMDetection framework.}
  \label{tab:mAPcoco}
\end{table}

\begin{table}
\setlength\tabcolsep{8pt}
\centering
\resizebox{\columnwidth}{!}{
\begin{tabular}{l|cccc}
\toprule
\multicolumn{1}{r|}{Domains:} & \multicolumn{1}{c}{Mobile} & \multicolumn{1}{c}{Web} & OS  & Avg.\\ 
\hline
Local Region                   &    38.4\%                   &              39.4\%            &         10.0\%    &   29.3\%   \\
Global Region                 &         45.6\%                    &            35.8\%              &       51.0\%    &    44.1\%    \\
\bottomrule
\end{tabular}}

\caption{Accuracy of the generated local region and global regions from our trained GUI region detection model compared with human annotation with IoU@0.5. Since the local regions are small, it brings a great challenge.}
\label{tab: ablation_bbox}
\end{table}

\paragraph{Inference Examples and Evaluation Results} After training, we utilize the GUI region detection model on unseen OS, Web, and mobile screenshots in our ScreenPR benchmark. We show four examples of the detection results in Figure \ref{fig:roc}. The identified local regions are marked by green rectangles, and the global regions are highlighted in red. 
Additionally, we manually annotate global regions for 20\% of randomly selected data from ScreenPR benchmark. With the human-annotated regions for the target points, we calculate the accuracy of our trained GUI region detection model as shown in Figure \ref{tab: ablation_bbox}. Trained on only mobile data, our model shows generalization ability to other unseen domains.

\begin{figure*}[hbt!]

\begin{subfigure}{.49\linewidth}
  \includegraphics[width=\linewidth]{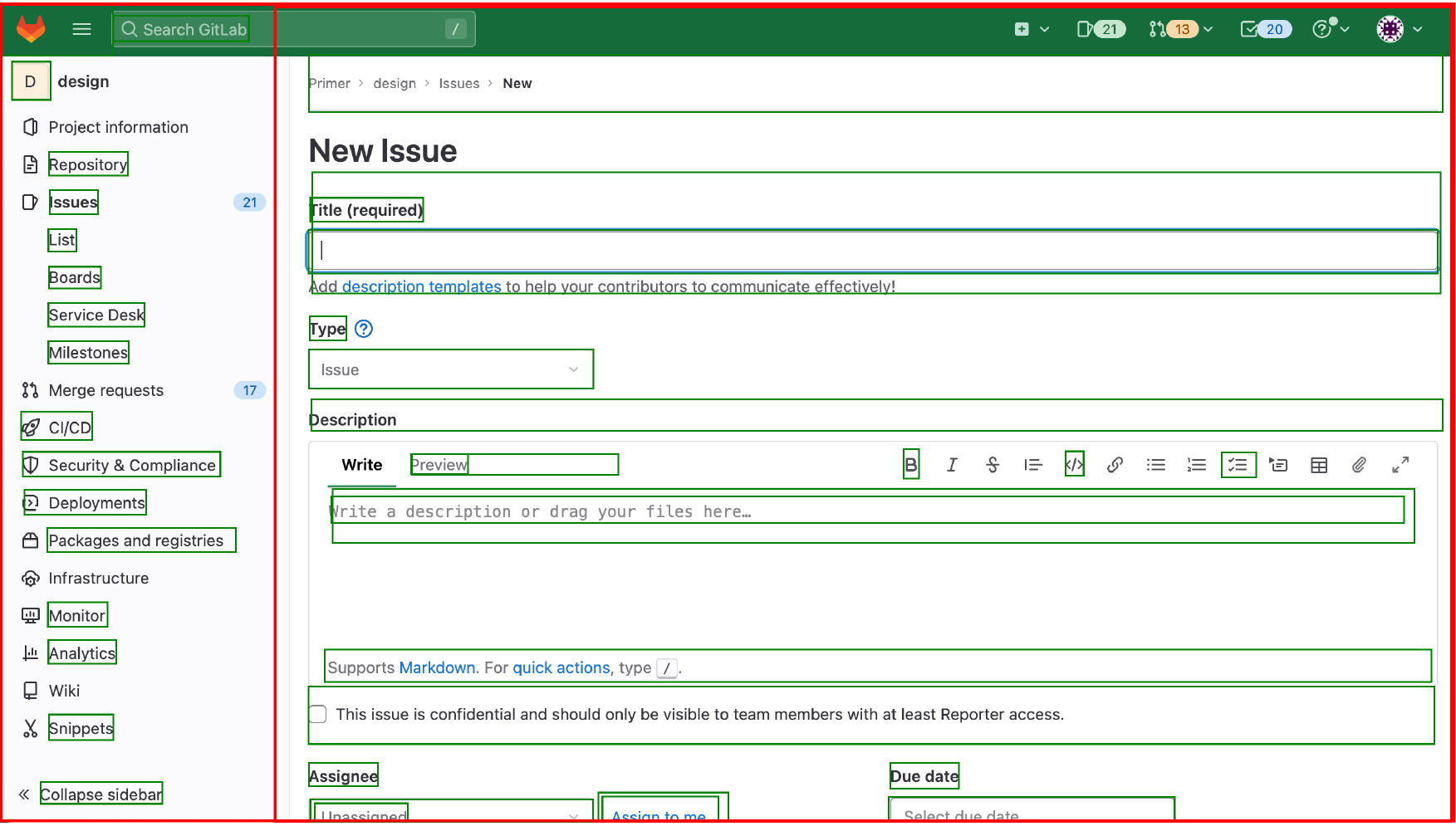}
  \caption{Web page}
  \label{demo:web}
\end{subfigure}\hfill
\begin{subfigure}{.49\linewidth}
  \includegraphics[width=\linewidth]{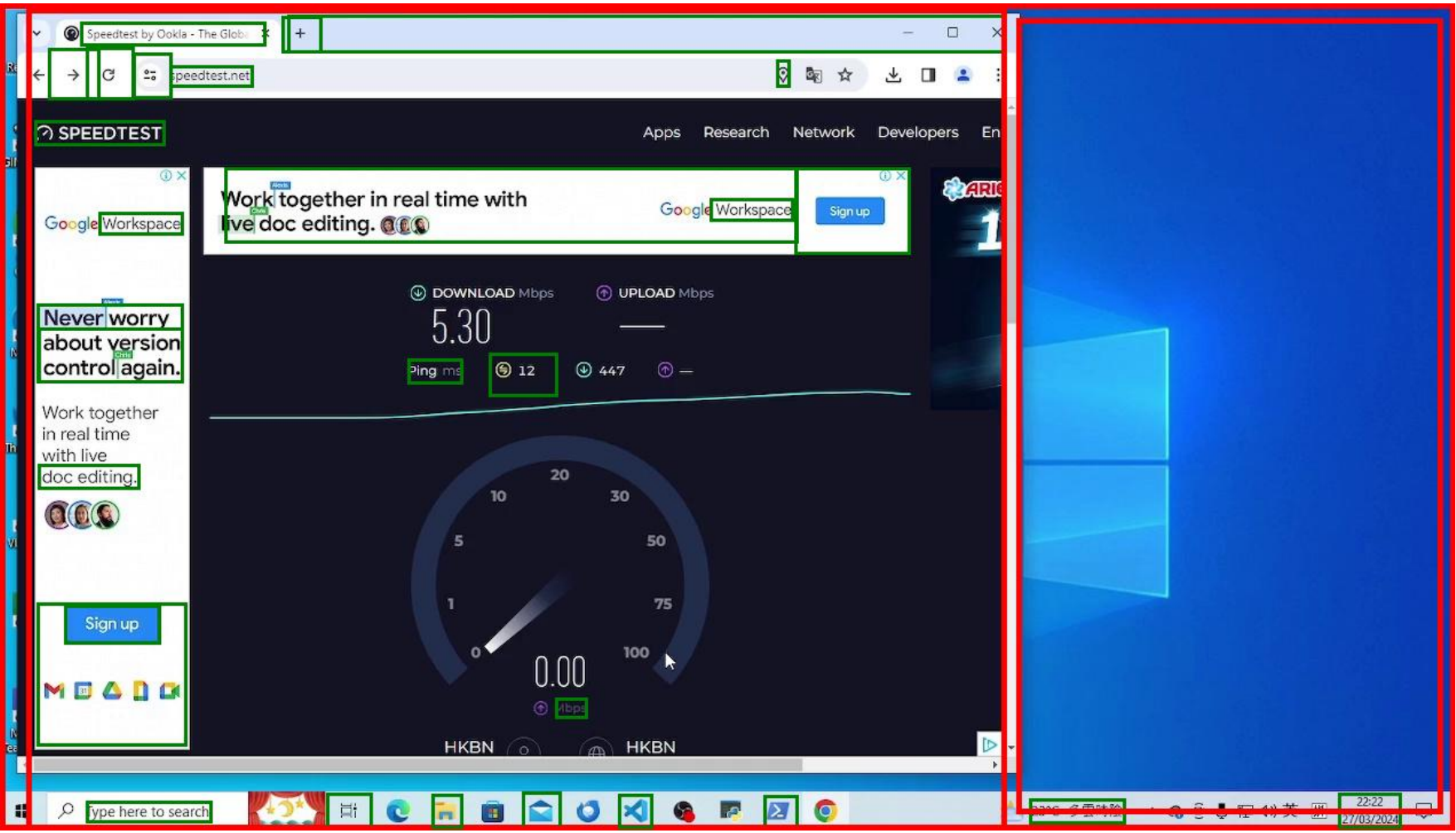}
  \caption{Windows GUI}
  \label{demo:pc_window}
\end{subfigure}

\medskip 
\begin{subfigure}{.63\linewidth}
  \includegraphics[width=\linewidth]{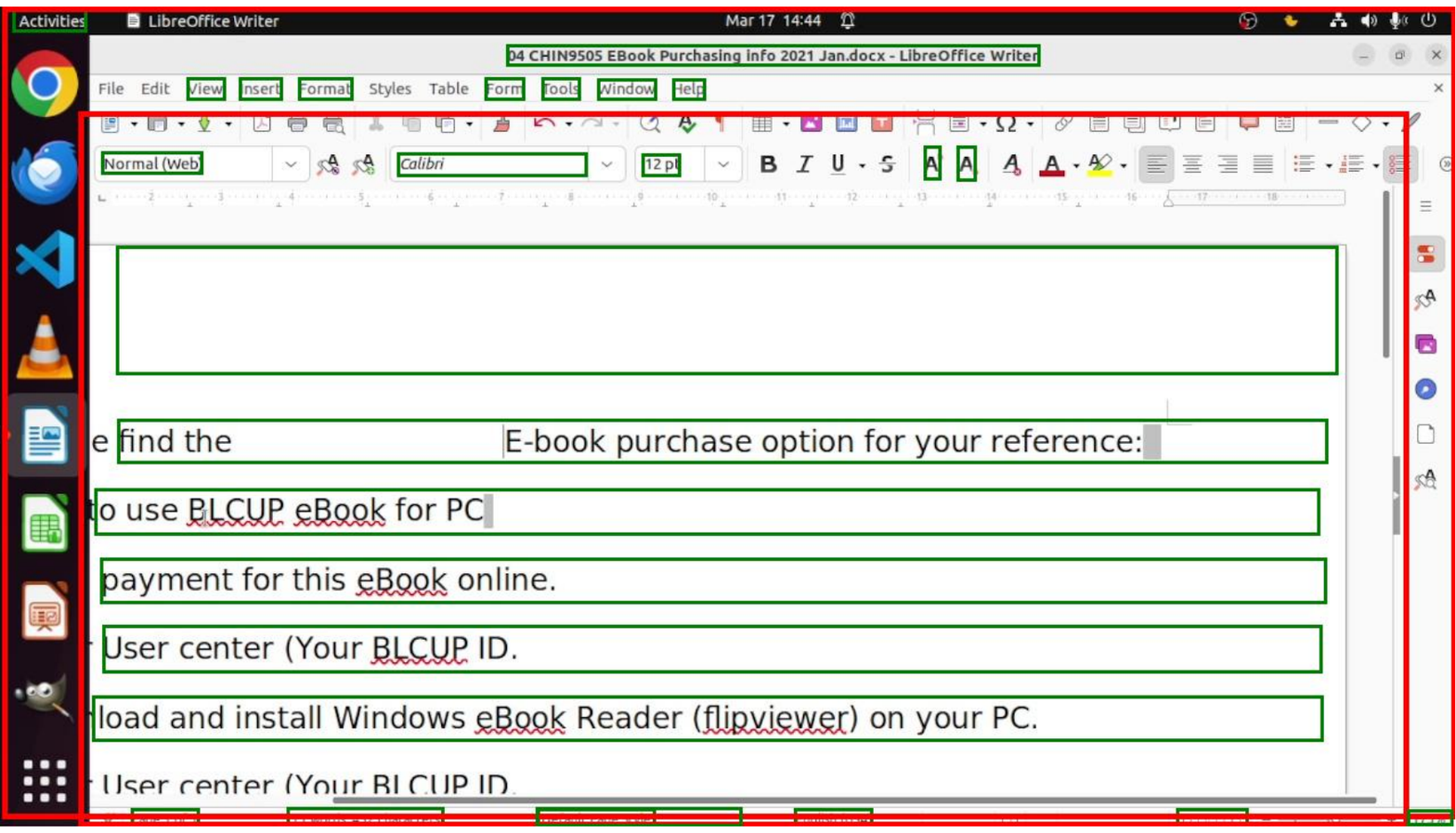}
  \caption{Ubuntu GUI}
  \label{demo:pc_ubuntu}
\end{subfigure}\hfill
\begin{subfigure}{.33\linewidth}
  \includegraphics[width=0.9\linewidth]{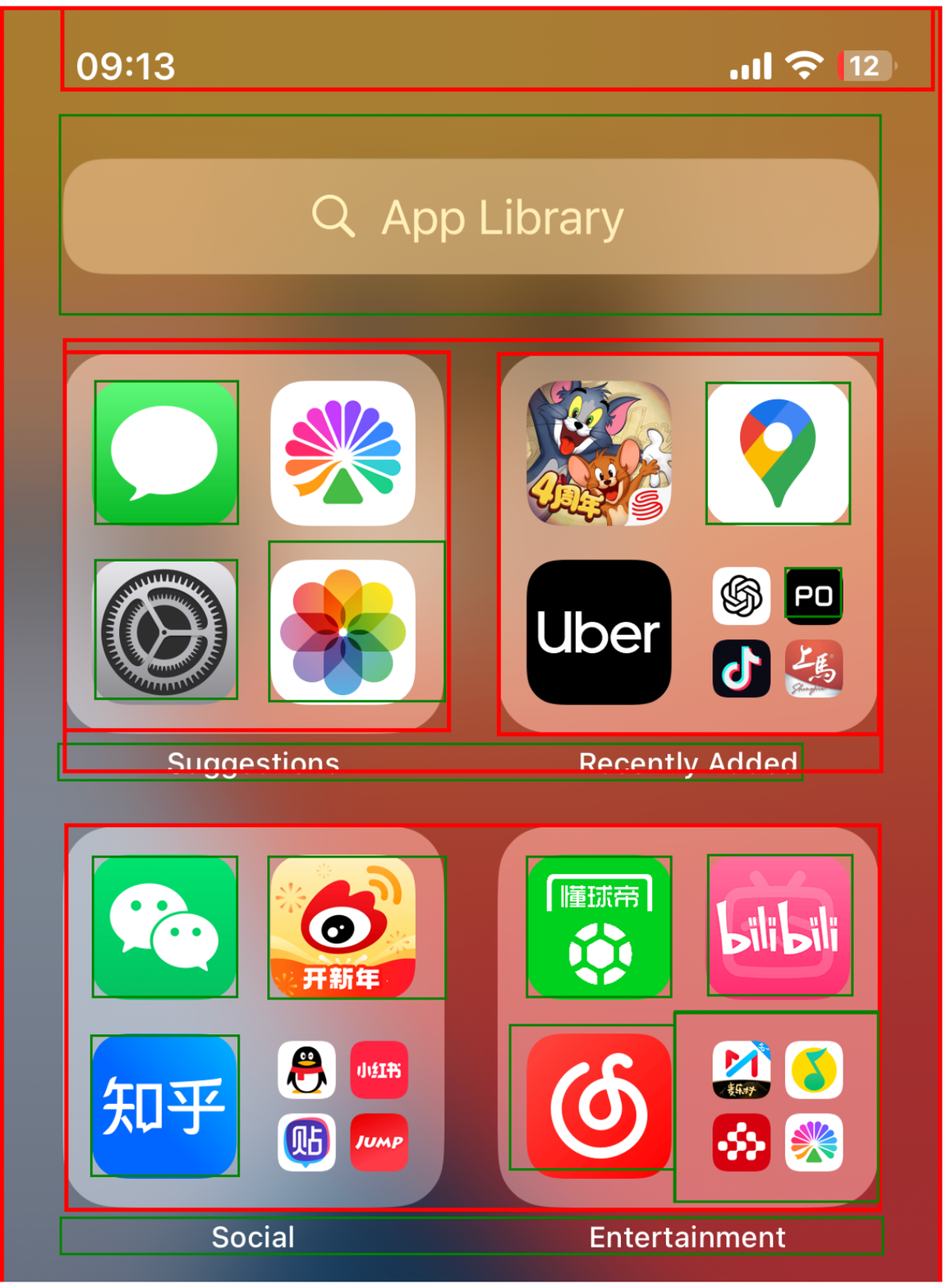}
  \caption{Unseen Mobile GUI}
  \label{demo:mobile}
\end{subfigure}

\caption{Inference results on Web, OS, and unseen Mobile GUI}
\label{fig:roc}
\end{figure*}

Except for some undetected ones, many local and global regions were successfully identified with promising confidential scores. We assume that GUI elements share some common design patterns learned by DINO and guarantee good performance for zero-shot inference on heterogeneous platforms.

\end{document}